\documentclass[10pt,journal,compsoc]{IEEEtran}

\usepackage{booktabs} 
\usepackage{subcaption}
\usepackage{multirow}
\usepackage{enumitem}
\usepackage{array}
\usepackage{bm}
\usepackage{amsthm}
\usepackage{amsmath}
\usepackage{amsfonts}
\usepackage{threeparttable}

\usepackage{graphicx}
\usepackage{algorithm}
\usepackage{algpseudocode}
\usepackage{verbatim}
\usepackage{amssymb}
 \usepackage{mathtools}
\usepackage{xcolor}
\usepackage{amsthm}
\usepackage{bbm}

\usepackage{float}
\usepackage{graphicx}
\usepackage{enumitem}
\usepackage{subcaption}
\usepackage{multirow}
\usepackage{CJKutf8}
\usepackage[table]{xcolor}
\usepackage{url} 
\usepackage{colortbl}
\definecolor{mygray}{gray}{.9}
\definecolor{mypink}{rgb}{.99,.91,.95}
\definecolor{mycyan}{cmyk}{.3,0,0,0}

\newcommand{\liang}{\color{black}}
\newcommand{\shijie}{\color{black}}
\newcommand{\zhang}{\color{black}}

\begin{document}

\title{Few-Shot Query Intent Detection via Relation-Aware Prompt Learning}

%
%

\author{Liang~Zhang,
        Yuan~Li,
        Shijie~Zhang,
        Zheng~Zhang,
        and Xitong~Li
\IEEEcompsocitemizethanks{
\IEEEcompsocthanksitem L. Zhang is with the Hong Kong University of Science and Technology (Guangzhou), China. Email: liangzhang@hkust-gz.edu.cn.
\IEEEcompsocthanksitem Y. Li and S. Zhang and X. Li are with Shenzhen MSU-BIT University, China. Email: roygodly123@gmail.com, zhang.shijie1101@gmail.com, lix@hec.fr.
\IEEEcompsocthanksitem Z. Zhang is with the Harbin Institute of Technology, China. Email: zhengzhang@hit.edu.cn.
\IEEEcompsocthanksitem Corresponding author: Xitong Li and Shijie Zhang}}

%
%

\markboth{Journal of \LaTeX\ Class Files,~Vol.~14, No.~8, August~2015}%
{Shell \MakeLowercase{\textit{et al.}}: Factor Graph based Collaborative Filtering}
%



\IEEEtitleabstractindextext{%
\begin{abstract}
Intent detection is a crucial component of modern conversational systems, {\liang since} accurately identifying user intent at the beginning of a conversation is essential for generating effective responses. 
{\liang Currently, most of the intent detectors can only work effectively with the assumption that high-quality labeled data is available (i.e., the collected data is labeled by domain experts).}
To ease this process, recent efforts have focused on studying this problem under a more challenging few-shot scenario. These approaches primarily leverage large-scale unlabeled dialogue text corpora to pretrain language models through various pretext tasks, followed by fine-tuning for intent detection with very limited annotations.
{\liang Despite the improvements achieved, existing methods have predominantly focused on textual data, neglecting to effectively capture the crucial structural information inherent in conversational systems, such as the \emph{query-query relation} and \emph{query-answer relation}. Specifically, the query-query relation captures the semantic relevance between two queries within the same session, reflecting the user’s refinement of her request, while the query-answer relation represents the conversational agent’s clarification and response to a user query.}
To address this gap, we propose SAID, a novel framework that integrates both textual and relational structure information in a unified manner for model pretraining for the first time. Firstly, we introduce a relation-aware prompt module, which employs learnable relation tokens as soft prompts, enabling the model to learn shared knowledge across multiple relations and become explicitly aware of how to interpret query text within the context of these relations. Secondly, we reformulate the few-shot intent detection problem using prompt learning by creating a new intent-specific relation-aware prompt, which incorporates intent-specific relation tokens alongside the semantic information embedded in intent names, helping the pretrained model effectively transfer the pretrained knowledge acquired from related relational perspectives. Building on this framework, we further propose a novel mechanism, the query-adaptive attention network (QueryAdapt), which operates at the relation token level by generating intent-specific relation tokens from well-learned query-query and query-answer relations explicitly, enabling more fine-grained knowledge transfer. Extensive experimental results on two real-world datasets demonstrate that SAID significantly outperforms state-of-the-art methods, achieving improvements of up to 27\% in the 3-shot setting. When equipped with the relation token–level QueryAdapt module, it yields additional performance gains of up to 21\% in the same setting. Experimental results also highlight the flexibility of both SAID and SAID (+QueryAdapt) as plug-and-play solutions compatible with various backbone models. 
\end{abstract}

\begin{IEEEkeywords}
Few-shot Intent Detection, Pretrained Language Model, Prompt Learning
\end{IEEEkeywords}}

\maketitle

\IEEEdisplaynontitleabstractindextext

%
\IEEEpeerreviewmaketitle

\section{Introduction}
\label{sec:intro}

Task-oriented conversational systems enable users to interact with computer applications through multi-turn natural language interactions to accomplish specific tasks with well-defined semantics. With recent advancements in large language models (LLMs) \cite{kenton2019bert,2020t5, brown2020language, touvron2023llama}, we have taken a step further and witnessed the emergence of modern open-domain conversational systems, known as LLM-powered {\liang chatbots such as ChatGPT} \cite{achiam2023gpt, team2023gemini, guo2025deepseek}. 
In this context, \emph{intent detection}, the task of classifying a {\liang user's behaviors (i.e., queries within each conversation)} into one of several mutually exclusive intent classes, becomes a cornerstone as accurately identifying user intent at the beginning of a conversation is crucial {\liang for downstream tasks such as effective response generation \cite{wu2023new}} and successful instruction execution \cite{qian2024tell}.  



{\liang However, to boost the performance of intent detection, costly and time-consuming manual annotation efforts are required.}
Additionally, user intents evolve over time, making it difficult to collect sufficient labeled data in a timely manner to cope with these dynamic intents in real-world commercial systems. These challenges highlight the need for sample-efficient methods that facilitate effective intent detection in a low-data scenario where only several labeled examples are available per intent (i.e., the so-called few-shot learning setups). Recently, this problem has garnered increasing research attention \cite{zhang2021effectiveness, shnarch2022cluster, zhang2022fine, zhang2021few}.

To address the few-shot intent detection problem, recent works have primarily leveraged large-scale pretrained language models (PLMs), such as BERT \cite{kenton2019bert} and RoBERTa \cite{liu2019roberta}. These models are usually pretrained on large-scale dialogue text corpora and then fine-tuned for intent detection utilizing different pretraining tasks. For example, IntentBERT \cite{zhang2021effectiveness} introduces supervised pretraining, which simultaneously optimizes a language modeling loss and a supervised loss on a small labeled dataset through a standard intent classification task. Shnarch et al. \cite{shnarch2022cluster} also adopt a classification task on top of BERT; however, instead of relying on manually labeled intent dataset, they generate{\shijie d} pseudo-labels through unsupervised clustering of query texts. LLM-Aug \cite{parikh2023exploring} further generates pseudo-labeled query texts by leveraging powerful large language models. In addition to task-oriented supervised pretraining, contrastive learning-based pretraining with data augmentation is also often employed with PLMs to tackle the few-shot intent detection problem. Specifically, CPFT \cite{zhang2021few} introduces a masked text augmentation strategy, treating a user query and its masked version as positive samples, and then pretrains a BERT encoder by bringing their representations closer through contrastive learning. Zhang et al. \cite{zhang2022fine} adopt a similar approach by introducing contrastive loss as a regularization technique to optimize the query representation space. However, they rely on a different data augmentation strategy borrowed from SimCSE \cite{gao2021simcse}, where the same query text with different dropout ratios is treated as positive samples. 



Nevertheless, most existing methods have primarily focused on pretraining PLMs based solely on the query text, overlooking the important structural information inherent in these conversational systems. Consider a typical example of a user interacting with an LLM-powered {\liang chatbot}, as illustrated in Figure \ref{fig:model}. In this scenario, Mary initiates a multi-turn query session related to the topic of ``climate change'', and the {\liang chatbot} generates a unified textual response for each specific query accordingly. From this example, two critical types of structural information can be identified: (1) the \emph{query-query relation}, which refers to the semantic relevance between two queries within the same session and reflects the user's refinement of her request, formed due to the continuity of user {\liang interaction} behaviors; and (2) the \emph{query-answer relation}, which reflects the LLM's clarification of a user query, emerging naturally during the interaction between the user and the {\liang chatbot}. Such structural information enhances the understanding of a given query from various relational perspectives and provides valuable insights for intent detection. Specifically, 
intent detection can be viewed as interpreting an incoming new query through the lens of intent name, which is essentially another relational perspective to clarify the query text and is closely related to both query-query and query-answer relations. If a pretrained model can learn to comprehend a user query from these typical relational perspectives already embedded in large-scale dialogue text corpora, it can generalize well and transfer this pretrained knowledge effectively, thereby facilitating the downstream few-shot intent detection task.


Motivated by this, we propose a new framework called SAID (\textbf{S}tructure-\textbf{A}ware pretraining for few-shot query \textbf{I}ntent \textbf{D}etection) in this paper, which unifies both textual and relational structure information inherent in conversational systems for model pretraining. The key idea is to introduce a relation-aware prompt module, which incorporates a few learnable relation tokens as special soft prompts, positioned between the original text tokens of a pair of queries or between a query and its answer. This module is highly flexible and enables the model to handle multiple relations jointly within a single framework. The PLM is then pretrained using the relation-aware prompts with a structure-aware masked language modeling task, encouraging it to learn shared knowledge across different relations and become explicitly aware of how to interpret query text within the context of these relations. When applied to the downstream few-shot intent detection task, intent-specific relation tokens are further introduced alongside the semantic information embedded in intent names, creating a new intent-specific relation-aware prompt that helps the model to effectively transfer the knowledge acquired from related relational perspectives during pretraining stage. 

In addition to the task reformulation via intent-specific relation-aware prompt at the modeling paradigm level, we further introduce a novel mechanism named QueryAdapt (\textbf{Query}-\textbf{Adapt}ive attention network) and propose an enhanced model named SAID (+QueryAdapt), to explicitly transfer relational knowledge from pretraining to downstream few-shot intent detection. QueryAdapt operates at the relation token level by generating intent-specific relation tokens from well-learned query-query and query-answer relations, enabling more fine-grained knowledge transfer. The main contributions of our paper are summarized as follows:

\begin{itemize}[leftmargin=*]
    \item Conceptually, we identify two important types of structural information inherent in conversational systems and examine the knowledge transfer across these relations in the context of intent detection, which leads to a new structure-aware problem formulation for the few-shot intent detection problem for the first time.
    
    \item Methodologically, we adopt a fundamentally different perspective from traditional contrastive learning-based methods and propose a novel pure prompt-based approach, which features in a flexible relation-aware soft prompt module that unifies query text and multiple relations within a single framework, along with a model-agnostic query-adaptive attention network that automatically generates soft prompts in an adaptive way.
    
    \item Empirically, extensive experimental results on two real-world datasets across four few-shot settings demonstrate that our SAID model significantly and consistently outperforms state-of-the-art baselines, achieving notable improvements of up to 27\% in the 3-shot setting on average. When equipped with the relation token–level QueryAdapt module, the enhanced version SAID (+QueryAdapt) yields an additional 21\% improvements over SAID in the same setting. Furthermore, we validate the flexibility of our framework by showing that both SAID and SAID (+QueryAdapt) can serve as plug-and-play solutions across various backbone architectures, while maintaining strong performance gains.
    
    \item {\liang Finally, we provide intent annotations for an unlabeled conversational dataset derived from the ChatGPT platform and make it publicly available to facilitate further research and development in this field.}
\end{itemize}

\section{Related Work}
\label{sec:related}

\subsection{Few-shot Intent Detection}
Pretraining on large-scale dialogue text corpora is a common strategy to address the data scarcity problem in few-shot intent detection task. Earlier efforts in this field are dedicated to continuously pretraining PLMs by leveraging self-supervised language modeling on large dialogue datasets and on the domain itself to tackle few-shot intent detection problem \cite{casanueva2020efficient, mehri2020dialoglue}. For example, CONVBERT \cite{mehri2020dialoglue} constructs an unlabeled dialogue corpus by concatenating user conversations into long text sequences and continues pretraining BERT upon this corpus, following standard BERT pretraining settings. Subsequent works extend this idea by designing additional pretraining tasks, incorporating either human-labeled intent datasets or pseudo-labeled query datasets \cite{zhang2021effectiveness, shnarch2022cluster, parikh2023exploring, zhang2023revisit}. For example, IntentBERT \cite{zhang2021effectiveness} finds that simultaneously optimizing a BERT-style language modeling loss alongside a supervised loss on a small human-labeled query intent dataset leads to better model performance. Shnarch et al. \cite{shnarch2022cluster} further demonstrate that unsupervised clustering of query texts can generate pseudo query labels, which can be used as valuable intermediate signals for model pretraining. Similarly, LLM-Aug \cite{parikh2023exploring} and DFT++ \cite{zhang2023revisit} generate pseudo-labeled or unlabeled augmented user query data by leveraging powerful large language models such as GPT-3.

Another group of works focuses on adapting contrastive learning techniques \cite{zhang2022region, chen2020simple} to this domain by designing various data augmentation strategies for user query data and pretraining PLMs using contrastive-based objectives. 
Specifically, Zhang et al. \cite{zhang2020discriminative} propose a data augmentation strategy that pretrains a model on annotated pairs from natural language inference (NLI) datasets and establishes a nearest neighbor classification schema for transfer learning, where two queries from the same intent class are labeled as positive. In addition to explicitly leveraging human-labeled data as augmentations, existing methods have also adopted heuristic and task-adaptive strategies, such as query text corruption \cite{zhang2021few} as well as different dropout masks \cite{zhang2022fine}. 

However, all these methods have primarily focused on pretraining PLMs using only textual data, failing to explicitly model the important structural information. Our work addresses this gap by identifying two crucial types of structural information and proposing the first structure-aware pretraining framework in the context of few-shot intent detection. 

\subsection{Prompting Method}
Most widely adopted PLMs, including BERT \cite{kenton2019bert}, T5 \cite{raffel2020exploring} and GPT \cite{radford2019language}, are pretrained using task agnostic language modeling objectives, which are often inconsistent with the specific goals of downstream tasks. To this end, prompt-based learning is proposed to reformulate these tasks as language modeling problems and facilitate PLMs to adapt effectively to these downstream tasks \cite{davison2019commonsense}. By defining new prompting functions, prompt-based learning even enables PLMs to perform few-shot and zero-shot learning, allowing them to adapt to new scenarios with limited labeled data \cite{gu2022ppt, nayaklearning}. Early methods in this field focus on manually crafting hard prompts, which are language instructions consisting of discrete text tokens prepended to the input text \cite{schick2021exploiting}. While effective, these hard prompts are challenging to optimize and rely heavily on domain-specific knowledge. To address the limitations of manual prompt design, recent works have proposed the use of soft prompts \cite{lester2021power, liu2022p}. Unlike hard prompts, which are composed of static tokens, soft prompts are continuous, learnable embeddings that can be optimized automatically via gradients during training. This approach has proven to be highly effective in various NLP tasks \cite{liu2023pre, sahoo2024systematic}. Aside from their success in the language domain, prompting methods have also been effectively applied in other areas including computer vision \cite{jia2022visual, bahng2022exploring} and graph neural networks \cite{liu2023graphprompt, sun2022gppt}. For instance, GPPT \cite{sun2022gppt} leverages prompting methods to bridge the gap between link prediction, which is used as a pretext task during pretraining, and the downstream node classification problem. 

Our work builds upon these advancements by incorporating soft prompt learning to enhance the few-shot intent detection task. By using relation-aware prompts in conjunction with a structure-aware masked language modeling task, we aim to improve the model's ability to capture and leverage the underlying relational structures embedded in large-scale dialogue text corpora, thereby facilitating better generalization and transfer learning in low-data scenarios. 

\section{Problem Statement}
\label{sec:pre}
In this section, we formulate the proposed problem of structure-aware pretraining for few-shot intent detection. Our problem relies on the existence of a large-scale unlabeled user dialogue dataset for model pretraining and a small labeled query intent dataset for model fine-tuning.

\noindent\textbf{Unlabeled Data.} Considering a LLM-powered conversational system where a user $u_j$ interacts with the LLM agent (i.e., chatbot) repeatedly for certain query purpose at time point $t$, which constitutes a query session denoted as $S_{u_j}^t=\{(q_i, a_i)|i=1,\dots,N\}$, where $N$ represents the length of this session, $q_i$ represents the user query and $a_i$ denotes a unified textual response generated by the LLM agent. {\liang 
Within each query session $S_{u_j}^t$, two important relations are introduced. Specifically, between two queries $q_i$ and $q_j$ within the same session $S_{u_j}^t$, the \emph{query-query relation}, denoted as $r_{qq}$ is established to reflect the semantic relevance between them. On the other hand, between a user query $q_i$ and its associated answer $a_i$, the \emph{query-answer relation}, denoted as $r_{qa}$, is also introduced to describe the correspondence between them. 
These critical relations are collectively referred to as structural information within the conversational systems in this paper.} 
By aggregating query sessions across different users and time points, we obtain a large-scale unlabeled user dialogue dataset $D_{unlabeled}=\{S_{u_j}^t|u_j \in \mathcal{U}, t_s \leq t \leq t_e\}$, where $\mathcal{U}$ represents sampled users and $[t_s, t_e]$ represents the observation window. 

\noindent\textbf{Labeled Data.} We are also given an intent dataset $D_{labeled}=\{(q_i, y_i)|i=1,\dots,M\}$, where $M$ represents the number of labeled queries, $y_i$ is the intent label of query $q_i$ and it is one of $C$ pre-defined intent labels denoted as $Y=\{y_i|i=1,\dots,C\}$. 

In our paper, we focus on a few-shot learning scenario. Following existing studies \cite{zhang2021few}, we assume that we have $K$ examples for each of the $C$ classes in our labeled data. In other words, we have $M=C*K$ training examples in total.

\noindent\textbf{Problem Definition.} Given the unlabeled user dialogue dataset $D_{unlabeled}$ and the labeled query intent dataset $D_{labeled}$, our problem is to learn an intent detection model $f_\theta(\cdot)$ that can accurately predict the intent of a new user query $q_m$.

Note that structural information is only utilized in the easily collected unlabeled dataset and during the model pretraining stage, rather than in the fine-tuning or inference stages. The inference procedure strictly follows the standard few-shot intent detection setup as adopted in prior work \cite{zhang2021few, zhang2021effectiveness, zhang2020discriminative}, meaning that intent is predicted based on the given query text, without relying on any historical information, to ensure both efficiency and fair comparison. This setup is also particularly important for identifying user intent at the beginning of a conversation, \textit{e.g.}, the first query in a new user session. Accurately recognizing user needs from the outset enables the system to generate higher-quality responses throughout the session, ultimately improving overall user satisfaction and engagement.


\begin{figure*}[th]
    \centering   \includegraphics[width=0.85\textwidth]{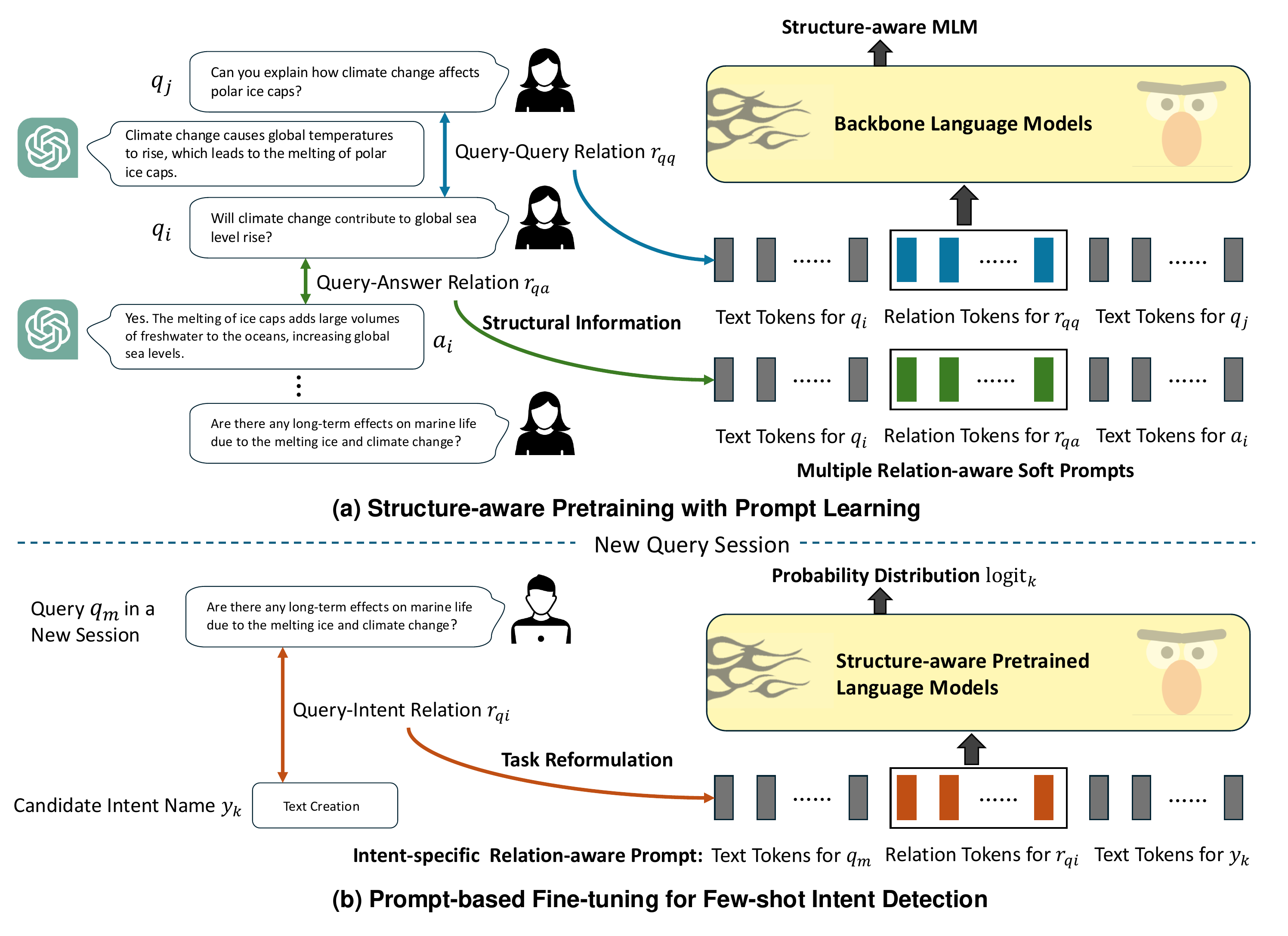}
    \vspace{-2mm}
    \caption{The model architecture of SAID. (a) Pretraining stage with relation-aware soft prompts; (b) Fine-tuning and inference stage with prompt-based task reformulation.}
    \vspace{-2mm}
    \label{fig:model}
\end{figure*}

\section{Methodology}
\label{sec:methodology}
In this section, we first present the technical details of SAID, with its overall architecture illustrated in Figure \ref{fig:model}. SAID comprises two key stages: a pretraining stage, where a PLM is pretrained to become structure-aware through self-supervised learning on large-scale unlabeled dialogue text corpora to enhance query understanding, and a subsequent supervised fine-tuning stage, where pretrained knowledge is transferred for few-shot intent detection using prompt-based task reformulation. 

In the pretraining stage, two important structures inherent in conversational systems that are often overlooked by existing works but would significantly enhance query understanding are identified, namely the \emph{query-query relation} and the \emph{query-answer relation}. These structures are unified with query text through soft prompt learning, resulting in a relation-aware prompt design that is easy for PLMs to interpret and flexible enough to incorporate multiple relations within a single framework. Building upon this foundation, a PLM is pretrained using the constructed relation-aware prompt as input through a structure-aware masked language modeling task, which serves as a natural extension of the traditional MLM task within the context of conversational structures.

In the fine-tuning stage, to facilitate effective knowledge transfer, we reformulate the intent detection task by introducing intent-specific relation tokens alongside the semantic intent names. This creates a new intent-specific relation-aware prompt, motivated by the idea that the semantics of different relations can complement each other. The prompt is then utilized as input for the pretrained PLMs for intent detection.

Building on this framework, we further propose a novel mechanism named the query-adaptive attention network (QueryAdapt) to explicitly transfer relational knowledge from pretraining to fine-tuning stage by generating intent-specific relation tokens from well-learned query-query and query-answer relations. This approach leverages the semantic proximity of relational structures to further enhance model performance under limited supervision.

\subsection{Structure-aware Pretraining with Prompt Learning}

Due to the inherently goal-driven and often informal or noisy nature of user dialogues \cite{henderson2020convert, zhang2020dialogpt}, models pretrained on other text corpora (e.g., BERT) struggle to effectively handle user queries in conversational systems. To address the semantic gap, these PLMs are continuously pretrained using dialogue text corpora, with a primary focus on the textual components, to enable them adapt to the unique characteristics of user queries \cite{mehri2020dialoglue, wu2020tod, zhang2021few, zhang2021effectiveness}. However, apart from the essential textual information, the multi-turn structure is also a unique characteristic of queries in conversational systems, as users commonly refine their queries through iterative exchanges on these platforms. Such multi-turn structure can be understood from two fundamental relational perspectives: the \emph{query-query relation} and the \emph{query-answer relation} as introduced in Section \ref{sec:pre}. 

\subsubsection{Relation Overview} On the one hand, the \emph{query-query relation}, denoted as $r_{qq}$, occurs between two queries posted by the user within the same session, reflecting how the user refines her requests and helping to interpret the user query from a human refinement perspective. On the other hand, the \emph{query-answer relation}, denoted as $r_{qa}$, describes the correspondence between a user query and its answer, illustrating how the model (often an LLM in modern conversational systems) responds to the user’s query and helping to interpret the user query from a model clarification perspective. They collectively enhance the understanding of a given query from various relational perspectives, providing valuable insights for interpreting a user query from the intent perspective, as the associated knowledge among them is highly relevant.

\subsubsection{Relation-aware Prompt Learning} To enable PLMs to be explicitly aware of these structures beneficial for query understanding while maintaining their ability to address the semantic gap between user query text and other common text through pretraining on dialogue text corpora, we propose to unify these two aspects (i.e., relations and text) together for model pretraining and adopt novel soft prompt learning to implement this idea. This represents the first attempt in this field to the best of our knowledge.
Specifically, we introduce relation tokens which function as specialized soft prompts and propose a relation-aware prompt that is easy for PLMs to interpret and helps them grasp knowledge about the unique textual characteristics of queries and the multi-turn structures inherent in conversational systems simultaneously.


Given a multi-turn query session $S_{u_j}^t$ initiated by user $u_j$ at time point $t$, we construct two relation-aware prompts for each query $q_i \in S_{u_j}^t$, corresponding to the two fundamental relations identified above as follows:
\begin{equation}
\label{eq:pretrain_prompt}
\begin{aligned}
	\text{Prompt}_{qq} &= \{q_i \ ; \ \mathbf{z}_{{qq}}^{(1)}, \cdots, \mathbf{z}_{{qq}}^{(m)} \ ; \ q_j\}, \\
    \text{Prompt}_{qa} &= \{q_i \ ; \ \mathbf{z}_{{qa}}^{(1)}, \cdots, \mathbf{z}_{{qa}}^{(m)} \ ; \ a_i\},
\end{aligned}
\end{equation}
where $\{\cdot ; \cdot\}$ is token concatenation operation, $\{ \mathbf{z}_{qq}^{(1)}, \cdots, \mathbf{z}_{qq}^{(m)} \}$ represent the $m$ relation tokens introduced for the query-query relation $r_{qq}$, and each $\mathbf{z}_{qq}^{(m)}$ is a $k$-dimensional trainable embedding vector; similarly, $\{ \mathbf{z}_{qa}^{(1)}, \cdots, \mathbf{z}_{qa}^{(m)} \}$ denote the corresponding relation tokens for the query-answer relation $r_{qa}$. These relation tokens are designed to act as specialized soft prompts in our framework. $q_j \in S_{u_j}^t$ is another user query in the same session as $q_i$, and $a_i$ refers to the response generated by the platform for query $q_i$. This prompt design integrates both relations and text that are essential for query understanding into a single framework. Note that our method is flexible and can be easily extended to accommodate more relations, such as similar queries explicitly labeled by humans.

\subsubsection{Structure-aware Masked Language Modeling} For each query in the unlabeled dialogue text corpora, 
we generate two corresponding relation-aware prompts, resulting in our structure-aware pretraining dataset. A PLM is then pretrained on this mixed dataset using the masked language modeling loss to grasp knowledge about the textual characteristics of queries and the multi-turn structures as follows: 
\begin{equation}
\mathcal{L} = - \frac{1}{M} \sum_{i=1}^{M} \log P(x_i | X_{\backslash i}),
\end{equation}
where $M$ is the number of masked tokens, $x_i$ is the text token at position $i$, $X_{\backslash i}$ represents the input sequence with the $i$-th text token masked. This pretraining step serves as a natural extension of the traditional MLM task in BERT, which relies solely on text data, by incorporating the proposed relation-aware prompts as inputs. It enables the model to learn how to interpret query text within the context of various relations.

\subsection{Few-shot Intent Detection}
\subsubsection{Implicit Knowledge Transfer via Task Reformulation}
Through self-supervised learning in the first stage, the model efficiently leverages a large number of unlabeled user queries alongside conversational structures. In the second stage, the model undergoes further tuning in a supervised manner using limited examples for few-shot intent detection. The fine-tuning design of our model is guided by the intuition that query intent detection can be framed as a query understanding task from the perspective of intent relation, which closely relates to both query-query and query-answer relations already exposed to and well captured by the PLMs during the pretraining phase. Therefore, to facilitate effective knowledge transfer, we propose to first reformulate the intent detection task from the relational perspective. 

Specifically, given a new user query $q_m$, we introduce the intent name as semantic information and build a new relation-aware prompt for each intent class as follows:
\begin{equation}
\label{eq:finetune_prompt}
\begin{aligned}
	\text{Prompt}_{qi} &= \{q_m \ ; \ \mathbf{z}_{{qi}}^{(1)}, \cdots, \mathbf{z}_{{qi}}^{(m)} \ ; \ y_k\}, \text{where} \ k=1,\dots,C. 
\end{aligned}
\end{equation}
Here, $\{ \mathbf{z}_{qi}^{(1)}, \cdots, \mathbf{z}_{qi}^{(m)} \}$ are new relation tokens introduced specifically for the query-intent relation (\textit{i.e.}, the relation $r_{qi}$ between $q_m$ and $y_k$ shown in Figure~\ref{fig:model}), $y_k$ is the intent name for $k$-th intent class. 

Instead of relying solely on user query text (\textit{i.e.}, $q_m$) as done in existing works, we utilize the intent-specific prompt as input to the pretrained PLMs for classification. The probability distribution of classifying $q_m$ into each intent class is calculated as follows:
\begin{equation}
\label{eq:classify}
\begin{aligned}
    \text{logit}_k^{m} &= \text{MLP}\left( f_{plm}\left( \text{Prompt}_{qi} \right) \right),\\
	\mathbf{p}_{m} &= \text{Softmax}\left( \left[\text{logit}_1^{m}, \cdots, \text{logit}_C^{m} \right] \right), 
\end{aligned}
\end{equation}
where $f_{plm}(\cdot)$ denotes the pretrained PLM. This model is then trained on the few-shot examples using cross-entropy loss. During inference, we predict the intent with the highest probability as the final output. We refer to this model as \textbf{SAID}, which also serves as our default model.

\begin{figure}[th]
    \centering   \includegraphics[width=0.90\linewidth]{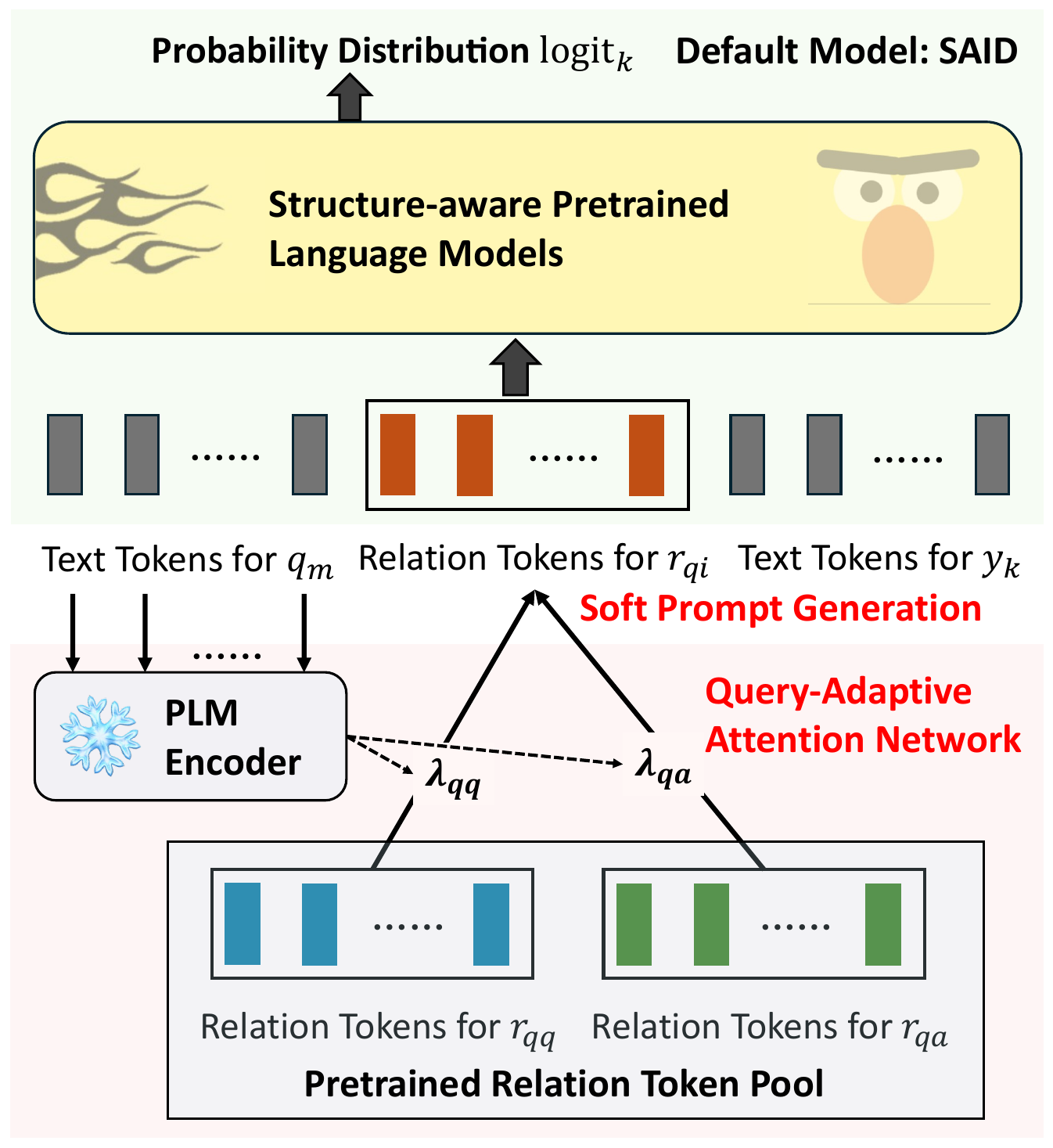}
    \vspace{-2mm}
    \caption{The model architecture of SAID (+QueryAdapt).}
    \vspace{-2mm}
    \label{fig:extension}
\end{figure}

\subsubsection{Explicit Knowledge Transfer via Soft Prompt Generation} By reformulating query intent detection from a relational perspective, we improve its alignment with the pretraining task, thereby facilitating the knowledge transfer of the overall modeling framework at the modeling paradigm level. However, the newly introduced intent-specific relation tokens are randomly initialized and may not be effectively optimized during the fine-tuning process with limited few-shot examples. To further enhance the knowledge transfer from the pretraining stage to the fine-tuning stage, we propose to transfer relation tokens explicitly at the fine-grained relation token level. The core intuition is that the query-intent relation is semantically closely related to the pretrained query-query and query-answer relations, allowing it to borrow knowledge from these well-learned relations. Specifically, we enable intent-specific relation tokens to be adaptively generated from the query-query and query-answer relation tokens, which have been effectively learned during the model pretraining phase. 

Formally, given a new user query  $q_m$, the corresponding query-intent relation tokens $\{ \mathbf{z}_{qi}^{(1)}, \cdots, \mathbf{z}_{qi}^{(m)} \}$ are generated by aggregating information from the pretrained query-query relation tokens $\{ \mathbf{z}_{qq}^{(1)}, \cdots, \mathbf{z}_{qq}^{(m)} \}$ and query-answer relation tokens $\{ \mathbf{z}_{qa}^{(1)}, \cdots, \mathbf{z}_{qa}^{(m)} \}$ using a query-adaptive attention network as follows:
\begin{equation}
\label{eq:relation_token_att}
\begin{aligned}
\mathbf{z}_{qi}^{(m)} &= \lambda_{qq} \cdot \mathbf{z}_{qq}^{(m)} + \lambda_{qa} \cdot \mathbf{z}_{qa}^{(m)}, \\
\left[ \lambda_{qq}, \lambda_{qa} \right] &= \text{Softmax}\left(\text{MLP}\left( f_{plm}\left(q_m \right) \right) \right),
\end{aligned}
\end{equation}
where $\lambda_{qq}$ and $\lambda_{qa}$ are the weights for query-query and query-answer relations respectively, generated based on the given user query $q_m$ encoded by the pretrained PLM $f_{plm}(\cdot)$. The intuition behind our model design is that different user queries may rely on these relations to varying extents when inferring user intent. For instance, longer and noisier queries might benefit more from query-query relations, leveraging query refinement to more clearly elucidate their underlying intents.

Based on Equation (\ref{eq:relation_token_att}), we construct the intent-specific relation-aware prompt $\text{Prompt}_{qi}$ as introduced in Equation (\ref{eq:finetune_prompt}) using the generated relation tokens, and ultimately predict the intent of a given user query $q_m$ using Equation (\ref{eq:classify}). We name this advanced model as \textbf{SAID (+QueryAdapt)}, and an overview of its framework is provided in Fig. \ref{fig:extension} (with the newly added soft prompt generation components highlighted in red).

\begin{table}[t]
 \caption{Representative examples in IntentChat and WildChat.}
\centering
\scalebox{0.57}{%
 \begin{tabular}{p{3.5cm}|p{2cm}|p{9cm}} 
 \toprule
 \midrule
 Query Intent & Dataset & Examples\\
 \hline
 \multirow{2}{*}{Text Creation}&IntentChat& \multicolumn{1}{>{\columncolor{mycyan}}l}{\begin{CJK}{UTF8}{gbsn}
写老人步履蹒跚的句子。
\end{CJK}}\\
 \cline{2-3}
 &WildChat&Write a letter of complaint to the local council. 140 words using the following prompts: recently you parked car/ returned/parking fine/no parking sign behind tree.\\
 \hline
 \hline
 \multirow{2}{*}{Knowledge Q \& A}&IntentChat
&\multicolumn{1}{>{\columncolor{mycyan}}l}{\begin{CJK}{UTF8}{gbsn}
红薯煮几分钟。
\end{CJK}}\\
\cline{2-3}
 &WildChat&While bioelectrochemistry deals with the electron transfer processes of biological systems, what types of biological systems are specifically studied and why?\\
 \hline
 \hline
 \multirow{2}{*}{Text Understanding}&IntentChat&\multicolumn{1}{>{\columncolor{mycyan}}l}{\begin{CJK}{UTF8}{gbsn}
你能分析一下这诗句的意境吗
\end{CJK}}\\
\cline{2-3}
 &WildChat&Summarize "Toto is a leading Japanese manufacturer of bathroom ceramic ware,
with annual worldwide sales of around \$5 bn. One of its best-selling
ranges is the Washlet lavatory, priced at up to \$5,000 and used in most
Japanese homes... \\
 \hline
 \hline
 \multirow{2}{*}{Mathematical Reasoning}&IntentChat&\multicolumn{1}{>{\columncolor{mycyan}}l}{\begin{CJK}{UTF8}{gbsn}
部分相干月牙形光束在海洋湍流中的交叉谱密度计算公式。
\end{CJK}}\\
\cline{2-3}
 &WildChat&An engineering system has two independent processes that occur one right after the other.  BOTH must be successful for the system to work.  Both systems work 90\% of the time.  What is the probability that the entire system will succeed?\\
 \hline
 \hline
 \multirow{2}{*}{Code}&IntentChat&\multicolumn{1}{>{\columncolor{mycyan}}l}{\begin{CJK}{UTF8}{gbsn}
WINCC.7.0如何批量替换。
\end{CJK}}\\
\cline{2-3}
 &WildChat&how to get the length of a {lol, lol2} list in python\\
 \hline
 \hline
 \multirow{2}{*}{Casual Conversation}&IntentChat&\multicolumn{1}{>{\columncolor{mycyan}}l}{\begin{CJK}{UTF8}{gbsn}
我在人海遇见你，却又转身不见你。
\end{CJK}}\\
\cline{2-3}
 &WildChat&why are you not responding. reply shortly.\\
 \hline
 \hline
 \multirow{2}{*}{Multimodal}&IntentChat&\multicolumn{1}{>{\columncolor{mycyan}}l}{---}\\
 \cline{2-3}
 &WildChat&As a prompt generator for a generative AI called "Midjourney", you will create image prompts for the AI to visualize. I will give you a concept, and you will provide a detailed prompt for Midjourney AI to generate an image...\\
 \hline
 \hline
 \multirow{2}{*}{Ambiguous Requests}&IntentChat
&\multicolumn{1}{>{\columncolor{mycyan}}l}{---}\\
\cline{2-3}
 &WildChat&you gpt4\\

 \midrule
   \bottomrule
 \end{tabular}}
 \label{tab:data_eg}
\end{table}

\begin{table}[h]
\caption{Prompt Description}
  \scalebox{0.58}{%
  \begin{tabular}{p{15cm}}
    \toprule
    \midrule
You are a diligent and responsible dataset annotator. Your task is to read the questions or requests input by users and select the most relevant label from eight new categories. The eight new categories are as follows: \\
1. **Text Creation**, such as creative writing, informative writing, evaluative writing, and editing, shortening, or expanding a text. The difference from text understanding is that text creation focuses on generating new content. \\
2. **Knowledge Q \& A**, such as explanatory questions, objective knowledge questions, opinion questions, and term searches. \\
3. **Text Understanding**, such as extracting information from a given text, semantic understanding, structural analysis, or language translation. Summarization belongs to text understanding, while shortening belongs to text creation. The distinction between summarization and shortening is that shortening needs to retain literary qualities. \\
4. **Mathematical Reasoning**, such as performing mathematical calculations, unit conversions, logical reasoning, or building models. \\
5. **Code**, such as programming problem-solving, code debugging, algorithm design, or data analysis. \\
6. **Casual Conversation**, such as everyday conversations, personal feelings and evaluations, role-playing, or sharing life experiences. \\
7. **Multimodal**, such as generating or understanding images, videos, audio, or describing image elements for text-to-image generation. \\
8. **Ambiguous Requests**, such as requests where the intention is unclear, hard-to-classify intentions, or long texts without a clear request. \\
The user input is: input\_text, and your output format should be in the JSON format: \\
 \qquad \qquad \qquad \qquad \qquad$\{$\\
 \qquad \qquad \qquad \qquad \qquad\qquad"summary": [summary is less than 80 words], \\
 \qquad \qquad \qquad \qquad \qquad \qquad"label": [label]\\
 \qquad \qquad \qquad \qquad \qquad$\}$\\ 
Label the intent category for the following data based on the above requirements. \\
$\{$input\_text$\}$ \\

    \midrule
   \bottomrule
\end{tabular}}

\label{tab:prompt}
\end{table}

\begin{table*}[h!]
	\renewcommand{\arraystretch}{1.05} 
	\centering 
	\caption{Results for intent detection across different few-shot settings. Tabular results are in percent. For each dataset, the first row reports the performance of strong zero-shot LLM baselines (such as LLaMA, Qwen, and Mistral) for reference. The best result is shown in boldface, and the second best result (excluding these LLM models) is underlined.}
    \large
	\resizebox{0.82\linewidth}{!}{
		\begin{tabular}{c|cc|cc|cc|cc}
			\toprule
			\multirow{3}{*}{Model} & \multicolumn{2}{c|}{{\textbf{3-shot}}} & \multicolumn{2}{c|}{{\textbf{5-shot}}} & \multicolumn{2}{c|}{{\textbf{10-shot}}} & \multicolumn{2}{c}{{\textbf{20-shot}}} \\
			\cmidrule{2-9}
			& Accuracy & F1 & Accuracy & F1 & Accuracy & F1 & Accuracy & F1 \\ 

            \bottomrule
            \toprule
            \multicolumn{9}{c}{\textbf{Dataset: IntentChat}} \\
            \midrule
            Llama-3.1-8B (zero-shot) & - & - & - & - & - & - &  \textit{25.08} & \textit{29.51} \\ 
            Qwen-7B (zero-shot) & - & - & - & - & - & - & \textit{72.57} & \textit{75.42} \\
            \midrule
		BERT & 34.98 & 41.34 & 33.06 & 39.76 & 43.29 & 46.85      & 57.53 & 61.58  \\
            RoBERTa & \underline{40.59} & \underline{47.46} & 43.60 & 48.19 & 51.57 & 55.99 & 62.78 & 67.47  \\ 
            DistilBERT & 31.10 & 36.63 & 43.24 & 47.43 & 52.01 & 56.96 & 56.24 & 60.90  \\
            ALBERT & 32.22 & 37.87 & 35.13 & 39.13 & 42.36 & 48.30& 44.17 & 50.33  \\
            CPFT & 39.77 & 40.04 & 41.92 & 40.35 & 35.00 & 38.81 & 43.39 & 43.68 \\
            IntentBERT & 37.60 & 42.19 & \underline{49.16} & \underline{55.03} & \underline{59.39} & \underline{64.34} & \underline{66.50} & \underline{70.83} \\
            DNNC & 29.99 & 43.49 & 40.86 & 44.66 & 49.21 & 55.35 & 58.31 & 64.65 \\
            SetFit & 40.04 & 40.78 & 45.45 & 47.58 & 54.24 & 57.74 & 60.69 & 64.70 \\
            LLM-Aug & 29.58	& 35.77 & 39.60	& 44.31 & 43.25	& 46.58 & 59.55	& 63.11  \\
            DFT++ & 20.09 &	16.24 & 22.19 & 16.99 &	29.27 & 23.64 &	38.60 & 29.19 \\
		\midrule
		SAID & \textbf{52.73} & \textbf{58.49} & \textbf{63.75} & \textbf{67.80} & \textbf{69.18} & \textbf{72.54} &	\textbf{74.05} & \textbf{76.44} \\
            Improvement & \textbf{29.91\%} & \textbf{23.24\%} & \textbf{29.68\%} & \textbf{23.21\%} & \textbf{16.48\%} & \textbf{12.74\%} & \textbf{11.35\%} & \textbf{7.92\%} \\

            \bottomrule
            \toprule
            \multicolumn{9}{c}{\textbf{Dataset: WildChat}} \\
            \midrule
            Llama-3.1-8B (zero-shot) & - & - & - & - & - & - &  \textit{52.01} & \textit{53.97} \\ 
            Mistral-7B (zero-shot) & - & - & - & - & - & - & \textit{50.94} & \textit{50.87} \\
            \midrule
            BERT & 22.88 & 21.96 & 35.44 & 36.95 & 40.13 & 43.00 & 49.16 & 53.57  \\
            RoBERTa & 33.87 & 35.61 & 41.85 & 45.05 & \underline{50.15} & \underline{54.29} & 53.48 & 57.72 \\ 
            DistilBERT & 28.35 & 30.32	& 34.30 & 36.28	& 38.96 & 41.93 & 45.93	& 50.50  \\
            ALBERT & 30.87 & 34.35 & 35.11 & 39.20 & 45.45 & 50.46 & 47.73 & 51.44  \\
            CPFT & 31.86 & 26.95 & 25.88 & 25.72 & 33.84 & 27.34 & 34.39 & 28.57   \\
            IntentBERT & 30.14 & 34.16 & \underline{44.41} & \underline{49.47} & 47.72 & 51.93 & 53.99 & 59.27  \\
            DNNC & 33.26 & 34.71 & 33.18 & 36.26 & 49.46 & 53.03 & \underline{55.98} & \underline{59.81}  \\
            SetFit & 30.82 & 33.71	& 38.73 & 42.24	& 46.28	& 51.13	& 52.34 & 57.30  \\
            LLM-Aug & 39.36	& \underline{42.78} & 42.18	& 45.60 & 45.13 & 47.75 & 50.76 & 53.94 \\
            DFT++ & \underline{40.23} &	30.60 & 41.19 & 35.80 & 42.85 & 36.95 & 49.63 & 40.12 \\
	    \midrule
		SAID & \textbf{50.03} & \textbf{53.05} & \textbf{51.20} & \textbf{54.73} & \textbf{57.17} &      \textbf{63.01} & \textbf{56.95} & \textbf{62.25} \\
            Improvement & \textbf{24.36\%} & \textbf{24.01\%} & \textbf{15.29\%} & \textbf{10.63\%} & \textbf{14.00\%} & \textbf{16.06\%} & \textbf{1.73\%} & \textbf{4.08\%} \\
            \bottomrule
	\end{tabular}}
	\label{tab:q1}
\end{table*}

\section{Experiments}
\label{section:experiment}
In this section, we conduct extensive experiments to answer the following important research questions regarding our default model SAID, and its enhanced variant SAID (+QueryAdapt). Specifically, for the default model SAID:
\begin{itemize}[leftmargin=*]
    \item \textbf{RQ1:} Can SAID achieve superior performance compared to state-of-the-art methods in intent detection accuracy across various few-shot settings?
    \item \textbf{RQ2:} Can SAID be applied as a plug-and-play framework to different PLM-based backbones, and how does it perform with these extensions?
    \item \textbf{RQ3:} How do different modules of SAID contribute to enhancing the overall model performance? 
    \item \textbf{RQ4:} How do different hyper-parameter settings impact the performance of SAID?
    \item \textbf{RQ5:} How efficient is SAID compared to state-of-the-art few-shot intent detection methods?
\end{itemize}

\noindent For the enhanced variant, SAID (+QueryAdapt): 
\begin{itemize}[leftmargin=*]
 \item \textbf{RQ6:} Does SAID (+QueryAdapt) outperform the default SAID model across various few-shot settings?
 \item \textbf{RQ7:} Can the proposed explicit knowledge transfer module via soft prompt generation (\textit{i.e.}, QueryAdapt) be applied as a general strategy across different PLM-based backbones to further enhance model performance over the default SAID framework?
 \item \textbf{RQ8:} What do the learned attention weights represent? Do they effectively capture the underlying characteristics of the query intent detection problem?
\end{itemize}

\begin{table*}[h]
	\renewcommand{\arraystretch}{1.05} 
	\centering 
	\caption{Results for plug-and-play experiments using three different PLM-based backbones. Tabular results are in percent.}
    \large
	\resizebox{0.82\linewidth}{!}{
		\begin{tabular}{c|cc|cc|cc|cc}
			\toprule
			\multirow{3}{*}{Model} & \multicolumn{2}{c|}{{\textbf{3-shot}}} & \multicolumn{2}{c|}{{\textbf{5-shot}}} & \multicolumn{2}{c|}{{\textbf{10-shot}}} & \multicolumn{2}{c}{{\textbf{20-shot}}} \\
			\cmidrule{2-9}
			& Accuracy & F1 & Accuracy & F1 & Accuracy & F1 & Accuracy & F1  \\ 
   
            \bottomrule
            \toprule
            \multicolumn{9}{c}{\textbf{Dataset: IntentChat}} \\
            \midrule
            
			BERT & 34.98 & 41.34 & 33.06 & 39.76 & 43.29 & 46.85 & 57.53 & 61.58\\
            SAID (BERT) & \textbf{52.73} & \textbf{58.49} & \textbf{63.75} & \textbf{67.80} & \textbf{69.18} & \textbf{72.54} & \textbf{74.05} & \textbf{76.44}  \\
            Improvement & \textbf{50.75\%} & \textbf{41.48\%} & \textbf{92.85\%} & \textbf{70.53\%} & \textbf{59.81\%} & \textbf{54.81\%} & \textbf{28.71\%} & \textbf{24.15\%} \\
            
   
            \midrule
            DistilBERT & 31.10 & 36.63 & 43.24 & 47.43 & 52.01 & 56.96 & 56.24 & 60.90  \\ 
            SAID (DistilBERT) & \textbf{50.17} & \textbf{53.41} & \textbf{56.24} & \textbf{60.38}	& \textbf{59.66} & \textbf{63.87} & \textbf{65.94} & \textbf{69.20} \\ 
            Improvement & \textbf{61.35\%} & \textbf{45.82\%} & \textbf{30.08\%} & \textbf{27.29\%} & \textbf{14.72\%} & \textbf{12.13\%} & \textbf{17.24\%} & \textbf{13.63\%}  \\

            \midrule
            ALBERT & 32.22 & 37.87 & 35.13 & 39.13 & 42.36 & 48.30& 44.17 & 50.33 \\
            SAID (ALBERT) & \textbf{49.51} & \textbf{50.36} & \textbf{50.62} & \textbf{54.83} & \textbf{61.27} & \textbf{64.22} & \textbf{57.32} & \textbf{61.32}  \\ 
            Improvement & \textbf{53.67\%} & \textbf{32.99\%} &	\textbf{44.08\%} & \textbf{40.13\%}	 & \textbf{44.66\%} & \textbf{32.95\%} & \textbf{29.78\%} & \textbf{21.84\%} \\
            
            \bottomrule
            \toprule
            \multicolumn{9}{c}{\textbf{Dataset: WildChat}} \\
            \midrule
            
			BERT & 22.88 & 21.96 & 35.44 & 36.95 & 40.13 & 43.00 & 49.16 & 53.57 \\
            SAID (BERT) & \textbf{50.03} & \textbf{53.05} & \textbf{51.20} & \textbf{54.73} & \textbf{57.17} & \textbf{63.01} & \textbf{56.95} & \textbf{62.25}  \\
            Improvement & \textbf{118.68\%} & \textbf{141.61\%} & \textbf{44.47\%} & \textbf{48.14\%} & \textbf{42.45\%} & \textbf{46.52\%} & \textbf{15.85\%} & \textbf{16.21\%} \\
            

            \midrule
            DistilBERT & 28.35 & 30.32	& 34.30 & 36.28	& 38.96 & 41.93	& 45.93	& 50.50  \\             
            SAID (DistilBERT) & \textbf{34.38} & \textbf{33.29} & \textbf{45.90} & \textbf{50.14} & \textbf{50.43} & \textbf{54.88} & \textbf{49.48} & \textbf{53.83} \\ 
            Improvement & \textbf{21.27\%} & \textbf{9.79\%}	& \textbf{33.83\%} & \textbf{38.21\%} & \textbf{29.44\%} & \textbf{30.88\%} & \textbf{7.74\%} & \textbf{6.59\%}	 \\

            \midrule
            ALBERT & 30.87 & 34.35 & 35.11 & 39.20 & 45.45 &	50.46 & 47.73 & 51.44 \\
            SAID (ALBERT) & \textbf{43.30} & \textbf{48.47} & \textbf{52.03} & \textbf{56.97}	& \textbf{58.75} & \textbf{62.89} & \textbf{59.03} & \textbf{64.20}  \\ 
            Improvement & \textbf{40.25\%} & \textbf{41.08\%} &	\textbf{48.20\%} & \textbf{45.35\%}	& \textbf{29.26\%} & \textbf{24.64\%} & \textbf{23.68\%} & \textbf{24.81\%} \\
            
            \bottomrule
	\end{tabular}}
	\label{tab:q2}
\end{table*}

\subsection{Experimental Settings}
\subsubsection{Datasets} Most publicly available datasets are typically limited to independent query data, lacking the structural information associated with query sessions. By collaborating with a LLM-powered technology company, we acquire user query data from its chatbot platform, covering the period from January 2024 to June 2024, resulting in the first dataset, named \textbf{IntentChat}. Additionally, we utilize a publicly available unlabeled user conversation dataset from a recent paper \cite{zhaowildchat}, which is the most relevant publicly available dataset in this domain to the best of our knowledge, though it only contains conversations from ChatGPT platform without any user intent annotations. We annotate this dataset with intent labels, creating the second dataset, named \textbf{WildChat}. This labeled dataset is made publicly available for the research community. The representative examples of each query intent are listed in Table~\ref{tab:data_eg}.



\noindent\textbf{IntentChat Dataset.} During the pretraining stage, we preprocess the sampled data by removing queries with zero length and filtering out sessions with fewer than three user interactions. This results in an unlabeled dataset consisting of 7,817 user query sessions, which includes a total of 91,595 query-answer pairs. We are also provided with an internally human-annotated dataset, from which we sample 20,000 queries to create our evaluation dataset. This dataset is used during the fine-tuning and testing stages. Following the experimental setup outlined by existing works \cite{zhang2021few}, we sample $K$-shot labeled queries per intent from the human-annotated dataset to fine-tune our model, while the remaining queries are regarded as the validation and test set. Note that we utilize user data with full consent obtained and in accordance with relevant privacy and ethical guidelines.


\noindent\textbf{WildChat Dataset.} For the pretraining dataset, we {\zhang first retain only the English conversations, and then }apply the same preprocessing steps as introduced in IntentChat, resulting in an unlabeled dataset consisting of 62,949 user query sessions, which include{\zhang s} a total of 354,104 query-answer pairs. For the evaluation dataset, we hold out and annotate 10,000 queries using GPT-4 Turbo. The prompt used for this labeling process is provided in Table \ref{tab:prompt}. To ensure annotation quality, we engaged two student experts to manually review the labeled queries, correct any identified errors, and ensure consistency in their opinions. 


\subsubsection{Evaluation Metrics} Accuracy and F1 are used to evaluate the effectiveness of the few-shot intent classification task following existing works \cite{zhang2022learn}. All experiments are repeated five times, and the average results are reported to ensure robustness of the evaluation.


\subsubsection{Baselines} We compare our approach against \textbf{thirteen} strong baseline models reported in the literature. Note that all methods adopt the same inference procedure and predict the intent of a given user query based on the input query text, thereby making it an inherently fair comparison.

The first category follows a classifier architecture that learns query representations using various PLM backbones, followed by a linear classification head optimized through cross-entropy loss. We include four commonly adopted PLMs in our experiments: \textbf{BERT} \cite{kenton2019bert}, \textbf{RoBERTa} \cite{liu2019roberta}, \textbf{DistilBERT} \cite{sanh2019distilbert} and \textbf{ALBERT} \cite{lan2019albert}. Specifically, BERT is a foundational transformer-based model that captures deep bidirectional representations, serving as the basis for numerous downstream NLP tasks. RoBERTa builds upon BERT by improving the pretraining approach through larger datasets and a dynamic masking strategy. DistilBERT , on the other hand, distills the knowledge of BERT into a more compact model, providing a balance between speed and accuracy while retaining the essential capabilities of the original BERT architecture. ALBERT is another widely adopted lightweight variant of BERT that reduces model size through parameter sharing and factorized embedding, achieving competitive performance with improved efficiency. 

The second category comprises methods specifically designed for the few-shot intent detection problem, including \textbf{CPFT} \cite{zhang2021few}, \textbf{IntentBERT} \cite{zhang2021effectiveness}, \textbf{DNNC} \cite{zhang2020discriminative} \textbf{SetFit} \cite{tunstall2022efficient}, \textbf{LLM-Aug} \cite{parikh2023exploring} and \textbf{DFT++} \cite{zhang2023revisit}. Specifically, CPFT is a two-stage pretraining framework for intent detection. In the first stage, the model learns using a self-supervised contrastive loss on a large set of unlabeled queries with data augmentations. In the second stage, it uses a supervised contrastive loss to consolidate query representations from the same intent class. IntentBERT pretrains BERT via supervised learning on a small human-labeled query intent dataset. DNNC is trained to identify the best-matched example from the training set through similarity matching. It enhances performance by incorporating data augmentations and natural language inference tasks during pretraining. SetFit is a Sentence Transformer model specifically designed for few-shot classification tasks. It uses triplet loss to fine-tune the transformer encoder and cross-entropy loss to optimize the classification head sequentially. LLM-Aug employs data augmentation to generate labeled data for intent detection by prompting large language models. Finally, DFT++ introduces a context augmentation method and leverages a sequential self-distillation technique to improve model performance.

Finally, the third category includes zero-shot large language model (LLM) baselines using the prompts shown in Table~\ref{tab:prompt}. For the IntentChat dataset, we include \textbf{LLaMA-3.1-8B} and \textbf{Qwen-7B}; for the WildChat dataset, we evaluate \textbf{LLaMA-3.1-8B} and \textbf{Mistral-7B}. Among them, LLaMA-3.1-8B is one of the most widely adopted general-purpose open-source LLMs. Qwen-7B is specifically optimized for Chinese understanding and generation, while Mistral-7B demonstrates strong performance on a wide range of English benchmarks. These models are approximately 60–70 times larger than BERT, our default backbone model. 

\subsubsection{Reproducibility} Our method is designed as a flexible, plug-and-play framework that can be seamlessly integrated with any PLM backbone. To balance effectiveness and efficiency, and in line with previous works~\cite{zhang2021few, zhang2021effectiveness} to ensure a fair comparison, we adopt the widely used BERT-base model as the default backbone for our experiments. In the pretraining stage, we pretrain our model for 4 epochs using the Adam optimizer with a learning rate of 3e-5. For the MLM loss, we follow the masking strategy used in BERT \cite{kenton2019bert}.
The number of relation tokens is set to 3. 
In the fine-tuning stage, we conduct experiments across four few-shot learning scenarios, specifically 3-shot (3 training examples per intent), 5-shot, 10-shot, and 20-shot settings. A hyperparameter search is performed for the learning rate over \{1e-5, 4e-5, 1e-4\}, using an early stopping strategy to prevent overfitting. Likewise, for the baseline methods, hyperparameters are carefully tuned according to their best performances on the validation set. All experiments are conducted on a single NVIDIA A100 GPU with 80GB of memory. 

\subsection{Performance Comparison (RQ1)}
The main results on two datasets are provided in Table \ref{tab:q1}, where the best performance is in boldface and the second best (excluding LLM models such as LLaMA, Qwen, and Mistral) is underlined, and the improvement of SAID over these second best methods is also presented in the last row. The zero-shot performance of the LLM models is included as a reference in the first row.

From these results, we can make the following observations. (1) The performance of RoBERTa is generally superior to that of BERT, which can be attributed to the fact that RoBERTa is trained with more robust optimizations and on larger datasets, providing it with stronger modeling capabilities compared to BERT. Similar findings can be observed by comparing BERT with DistilBERT where BERT has more parameters and exhibits more powerful capabilities. (2) The few-shot intent detection models generally achieve the best performances in most cases among all baselines. For instance, even with BERT as the backbone, IntentBERT, pretrained on large-scale dialogue text corpora, effectively adapts to the unique characteristics of user queries and bridges the semantic gap, achieving even better performance than RoBERTa. (3) Unlike these baselines that focus solely on text, we introduce additional structural information beneficial for query understanding during pretraining, enabling us to fully leverage the unlabeled dialogue text corpora. Meanwhile, we reformulate the few-shot intent detection problem from a relational perspective using prompt learning during fine-tuning, facilitating effective knowledge transfer from pretrained relations. As demonstrated in Table \ref{tab:q1}, these novel designs enable our model to consistently outperform all the baselines across all metrics on both datasets. And the average margin in terms of accuracy (F1 score) reaches 21.86\% (16.78\%) for the IntentChat dataset and 13.85\% (13.70\%) for the WildChat dataset, respectively. (4) Even with a small PLM backbone (BERT), our model SAID achieves comparable or even superior performance to LLMs that are 60–70 times larger, using very limited supervision (\textit{e.g.}, 5-shot labeled data on the WildChat dataset). This result underscores the effectiveness of our proposed relation-aware prompt-based pretraining and finetuning framework. Furthermore, equipped with an explicit knowledge transfer strategy based on soft prompt generation, our model can achieve better performance compared to these LLMs, while requiring fewer labeled examples. Additional results are provided in Section~\ref{sec:ex_rq6}. Given the substantial computational and financial costs associated with these LLMs, our approach offers a highly cost-effective, scalable, and better alternative for real-world deployment.

\subsection{Extension to Various Backbones (RQ2)} \label{sec:rq2}
To verify the plug-and-play capability and robustness of our framework, we replace the backbone model (BERT-base) with other commonly adopted PLMs, including both DistilBERT and ALBERT. The experiment results on both datasets are presented in Table \ref{tab:q2}, where the performance improvement of the SAID framework over each corresponding backbone model is presented. 
Despite variations in architecture and/or model size between BERT, DistilBERT, and ALBERT, our framework consistently delivers significant improvements across all PLMs in all few-shot settings. We also observe that, when equipped with SAID, previously underperforming models such as DistilBERT and ALBERT could outperform the best baseline in Table \ref{tab:q1}. For instance, SAID (DistilBERT) and SAID (ALBERT) achieve much better performance in the 5-shot setting on the WildChat dataset compared to IntentBERT (the strongest baseline excluding LLM models), demonstrating that SAID can effectively adapt to different PLMs while maintaining strong performance. Overall, these results validate the robustness and plug-and-play capability of our proposed framework.

\begin{table*}[h]
	\renewcommand{\arraystretch}{1.05} 
	\centering 
	\caption{Results for ablation study on IntentChat. Tabular results are in percent.}
    \large
	\resizebox{0.82\linewidth}{!}{
		\begin{tabular}{c|cc|cc|cc|cc}
			\toprule
			\multirow{3}{*}{Model} & \multicolumn{2}{c|}{{\textbf{3-shot}}} & \multicolumn{2}{c|}{{\textbf{5-shot}}} & \multicolumn{2}{c|}{{\textbf{10-shot}}} & \multicolumn{2}{c}{{\textbf{20-shot}}} \\
			\cmidrule{2-9}
			& Accuracy & F1 & Accuracy & F1 & Accuracy & F1 & Accuracy & F1 \\ \midrule
            w/o PT & 45.92 & 52.52 & 57.18 & 61.62	& 55.28 & 59.15	& 65.88 & 69.53 \\
			w/o REL & 46.51 & 52.09 & 56.54 & 61.80	& 59.82 & 64.46	& 67.30 & 71.14 \\
            w/o Q-A & 51.68 & 55.73 & \underline{58.85} & \underline{63.50} & 62.15 & 67.51 & 64.91 & 70.65 \\ 
            w/o Q-Q & \textbf{54.24} & \textbf{60.50} & 57.75 & 63.47 & \underline{66.63} & \underline{71.03} & \underline{69.72} & \underline{73.92} \\ 
            \midrule
            w/o PLFT & 41.26 & 46.85 & 43.72 & 49.15	& 55.37 & 60.06	& 64.48 & 68.85 \\
		    \midrule
            SAID & \underline{52.73} & \underline{58.49} & \textbf{63.75} & \textbf{67.80} & \textbf{69.18} & \textbf{72.54} & \textbf{74.05} & \textbf{76.44} \\
			\bottomrule
	\end{tabular}}
	\label{tab:q3}
\end{table*}

\subsection{Ablation Study (RQ3)}
To demonstrate how different components of the proposed framework affect the performance, we conduct ablation studies by removing or replacing key modeling modules from the full architecture. 
Specifically, we investigate five simplified versions of our model: (1) \textbf{w/o PT} removes the entire pretraining phase; (2) \textbf{w/o REL} excludes all structural information during pretraining, relying solely on unlabeled query text to pretrain BERT; (3) \textbf{w/o Q-A} and \textbf{w/o Q-Q} eliminate the query-answer relation and the query-query relation during pretraining, respectively; (4) Finally, \textbf{w/o PLFT} replaces the proposed prompt-learning based fine-tuning approach with a traditional classification head that relies solely on query text for intent classification (as done in existing works). The ablation results on the IntentChat dataset are presented in Table  \ref{tab:q3}. 

\noindent\textbf{Benefits of structure-aware pretraining.} From the results summarized in Table \ref{tab:q3}, we can make the following observations. (1) In most cases, the performance of w/o REL exceeds that of w/o PT, revealing that pretraining with unlabeled query text helps bridge the semantic gap and enables the PLM to adapt to the unique characteristics of user queries in conversational systems. (2) However, w/o REL is often outperformed by w/o Q-A and w/o Q-Q. For example, w/o Q-Q surpasses w/o REL by approximately 11.38\% and 10.19\% in terms of average accuracy and F1 score respectively in the 10-shot setting, showing that incorporating additional structural information during pretraining is crucial for better query understanding. (3) In general, the full model SAID significantly outperforms both w/o Q-A and w/o Q-Q, demonstrating the advantages of integrating various critical relations in conversational systems within a unified framework. Overall, these experiments collectively highlight the importance of our proposed structure-aware pretraining framework. 

\noindent\textbf{Benefits of prompt-learning based fine-tuning.} Compared to w/o PLFT, SAID achieves better performance in all metrics and across all few-shot settings, with an average improvement of 28.35\% in accuracy and 23.64\% in F1 score, respectively. These results verify the necessity of prompt-learning based task reformulation for effective knowledge transfer and highlight its advantages over traditional solutions.

\subsection{Parameter Analysis (RQ4)}
In this section, we investigate the impacts of two important hyperparameters, the masking ratio and the number of pretraining epochs, on SAID using the IntentChat dataset.

\begin{figure}[t!]
	\centering
	\begin{subfigure}[b]{0.235\textwidth}
		\includegraphics[width=\textwidth]{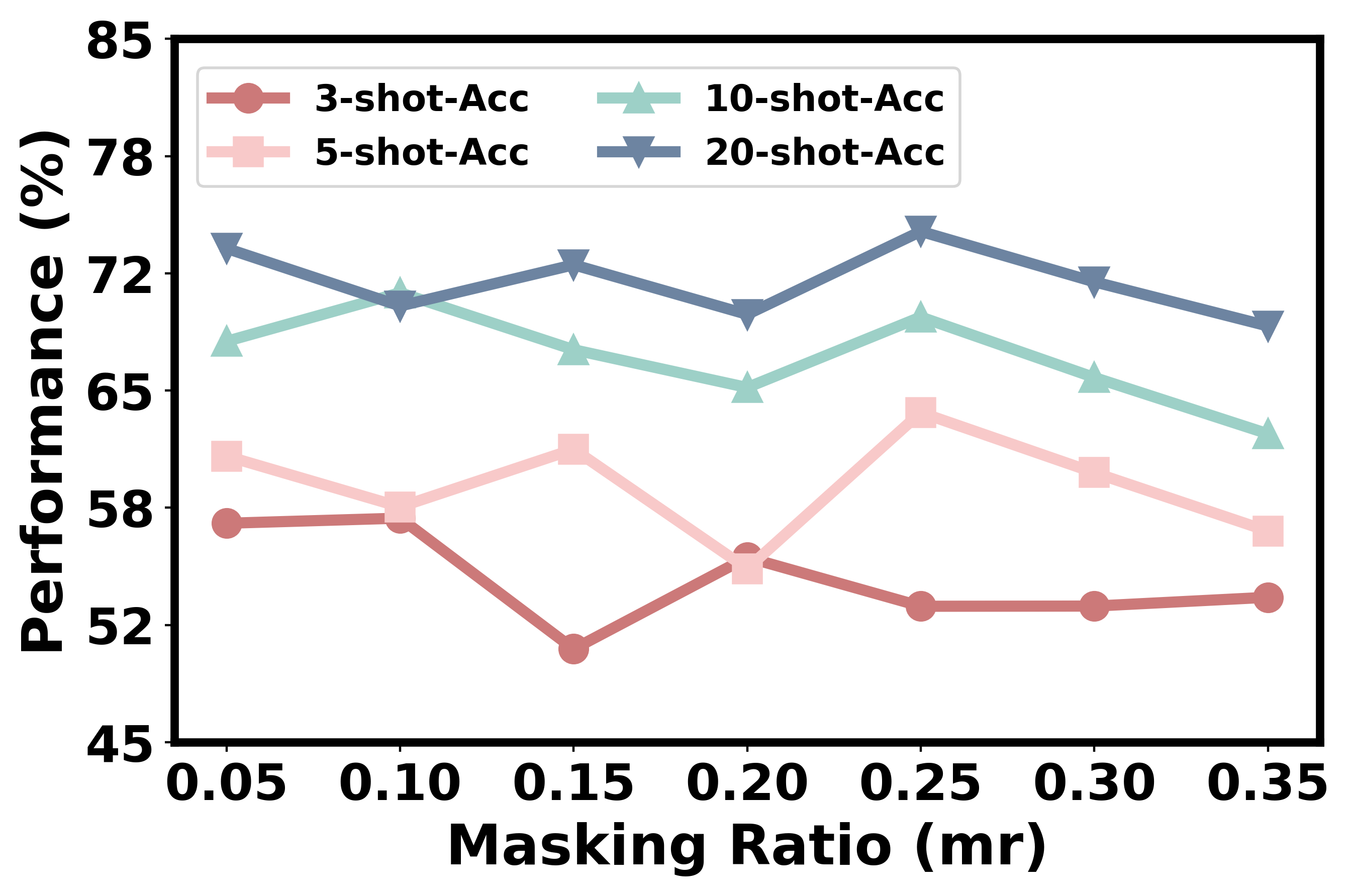}
		\caption{Accuracy}
		\label{fig:mask_acc}
	\end{subfigure}
	\begin{subfigure}[b]{0.235\textwidth}
		\includegraphics[width=\textwidth]{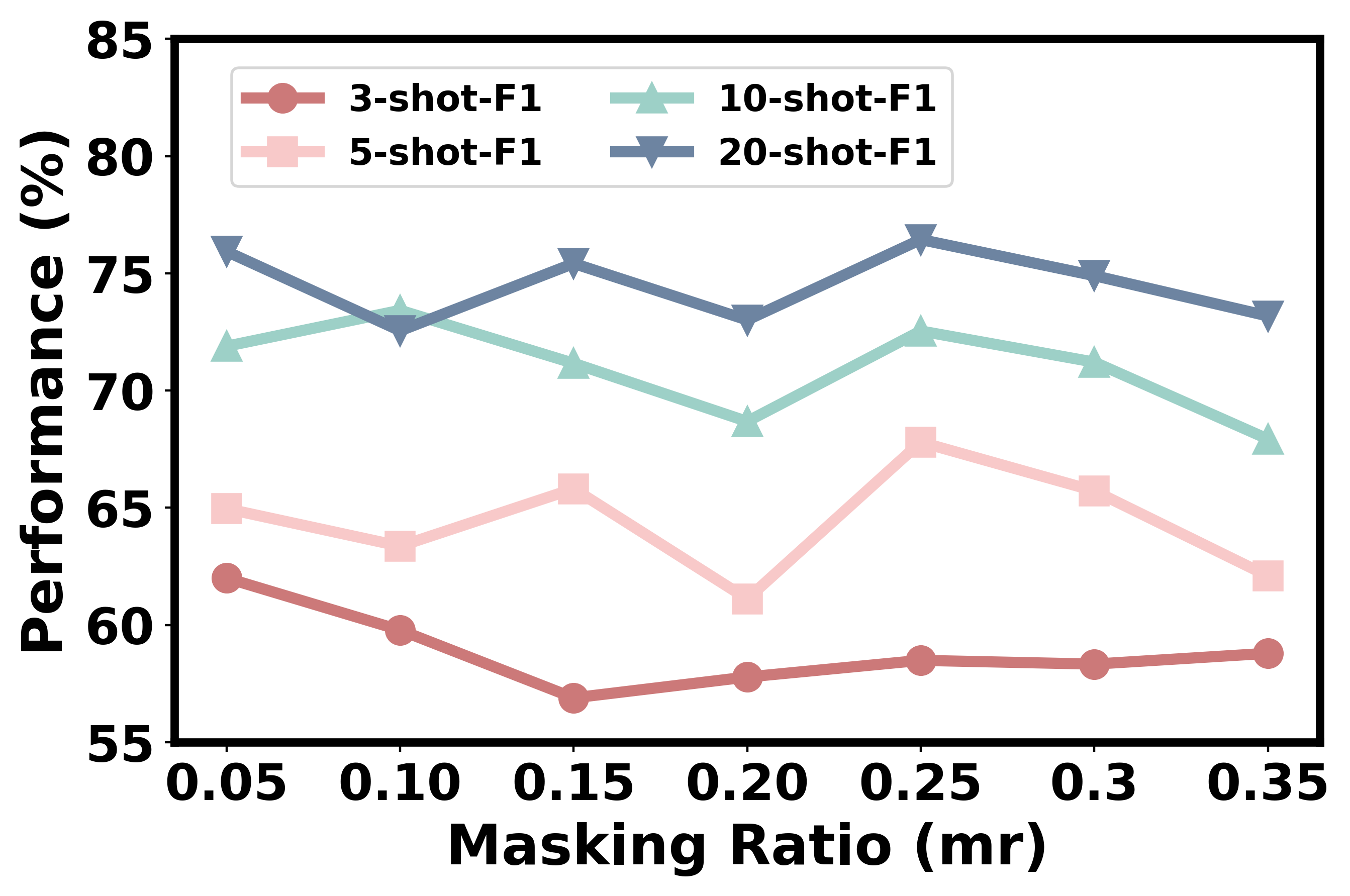}
		\caption{F1}
		\label{fig:mask_f1}
	\end{subfigure}
    \vspace{-3mm}
	\caption{Impact of masking ratio on IntentChat.}
	\label{fig:mask_ratio}
    \vspace{-3mm}
\end{figure}

\noindent\textbf{Impact of Masking Ratio.} As shown in Figure \ref{fig:mask_ratio}, the model performance initially declines as the masking ratio increases, but starts to improve after it exceeds 0.20. The best performance across most metrics is achieved at a masking ratio of $mr=0.25$, which is different from the optimal value of $mr=0.15$ found in BERT's standard masked language modeling task. This suggests that by introducing additional relation contexts, a more challenging pretraining setting is needed for improved representation learning.

\begin{figure}[t!]
	\centering
	\begin{subfigure}[b]{0.235\textwidth}
		\includegraphics[width=\textwidth]{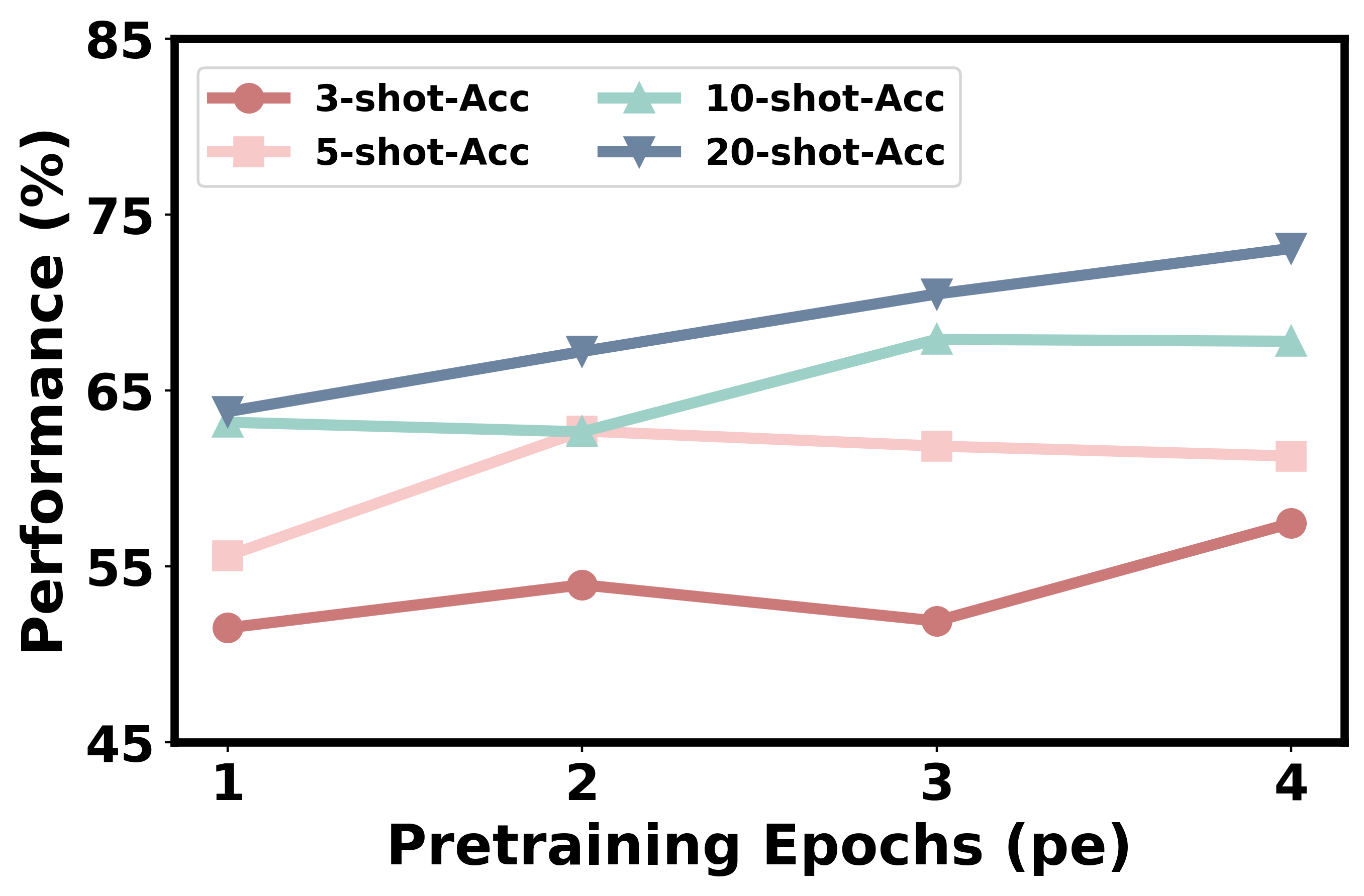}
		\caption{Accuracy}
		\label{fig:epoch_acc}
	\end{subfigure}
	\begin{subfigure}[b]{0.235\textwidth}
		\includegraphics[width=\textwidth]{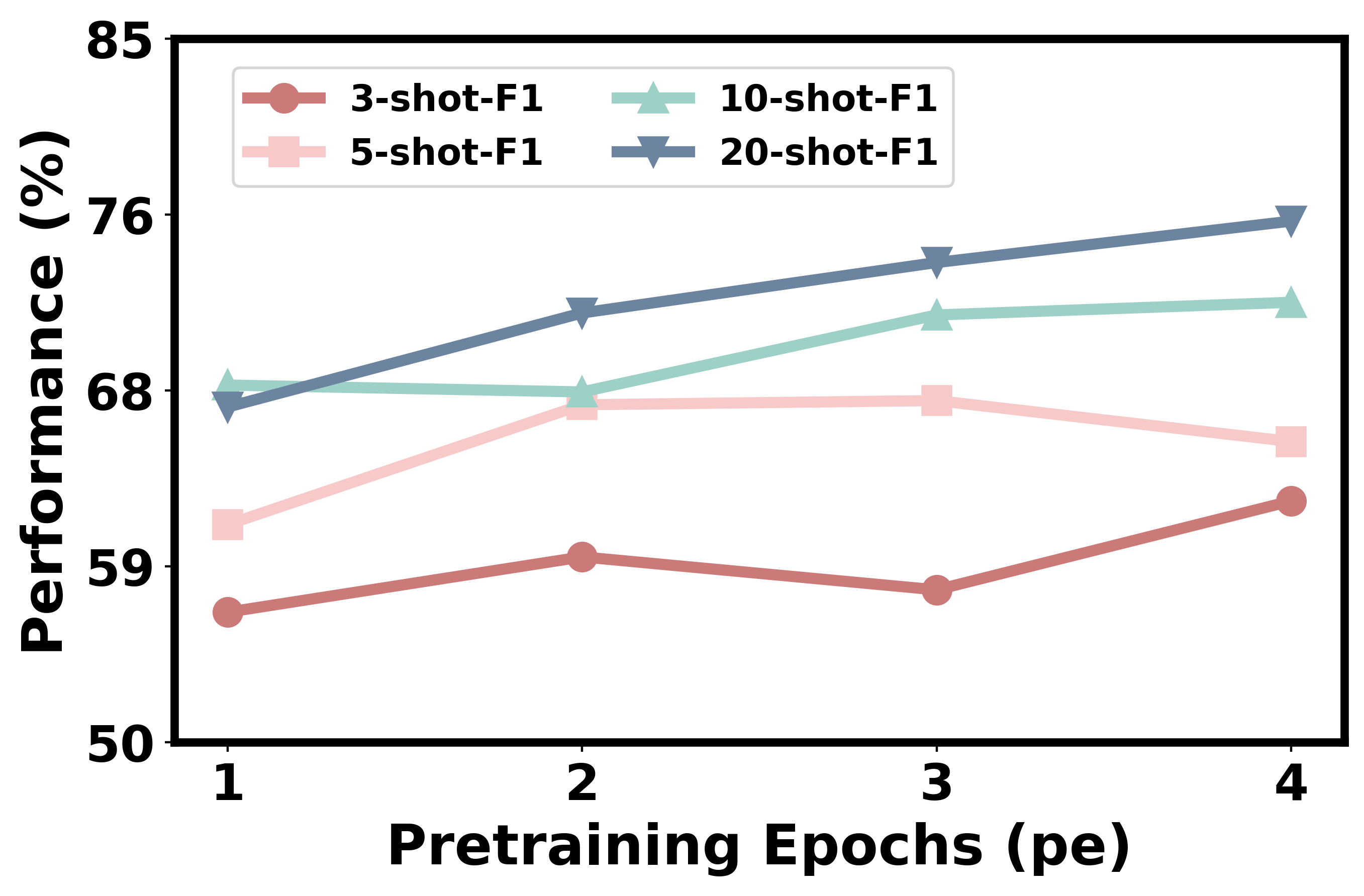}
		\caption{F1}
		\label{fig:epoch_f1}
	\end{subfigure}
    \vspace{-3mm}
	\caption{Impact of pretraining epochs on IntentChat.}
	\label{fig:pre_epochs}
    \vspace{-3mm}
\end{figure}

\noindent\textbf{Impact of Pretraining Epochs.} Figure \ref{fig:pre_epochs} illustrates the impact of pretraining epochs on the model performance. It's evident that, in most cases, model performance improves as the number of pretraining epochs increases. For example, the average accuracy in the 20-shot setting when 
$pe=4$ is over 10\% higher than when $pe=1$, which validates the benefits of our structure-aware pretraining from another perspective. 

\begin{table*}[h]
	\renewcommand{\arraystretch}{1.05} 
	\centering 
	\caption{Results for efficiency analysis on IntentChat.}
    \large
	\resizebox{0.85\linewidth}{!}{
		\begin{tabular}{c|cccc|c}
			\toprule
			\multirow{3}{*}{Model} & \multicolumn{4}{c|}{{\textbf{Average Training Time per Epoch (s)}}} & \textbf{Average Inference Time per Query (s)} \\
			\cmidrule{2-6}
			& 3-shot & 5-shot & 10-shot & 20-shot & N/A \\ \midrule
            CPFT & 2.61 & 2.71 & 2.87 & 3.24 & 2.6e-4 \\
            IntentBERT & 0.19 & 0.20 & 0.32 & 0.77 & 3.5e-4 \\
            DNNC & 5.59 & 12.09 & 40.78 & 149.53 & 0.1965 \\
            SetFit & 6.44 & 7.68 & 11.81 & 20.42 & 7.7e-4 \\
            LLM-Aug & 0.25 & 0.43 & 0.80 & 1.59 & 4.6e-4 \\
            DFT++ & 27.82 &	29.92 &	32.88 & 39.35 & 0.0080\\
		    \midrule
            SAID (BERT) & 2.25 & 3.51 & 7.14 & 15.38 & 3.3e-4 \\
            SAID (ALBERT) & 1.71 & 2.51 & 5.19 & 10.32 & 2.4e-4 \\
			\bottomrule
	\end{tabular}}
	\label{tab:effi}
\end{table*}

\begin{table*}[h!]
	\renewcommand{\arraystretch}{1.05} 
	\centering 
	\caption{Results of soft prompt generation for intent detection across various few-shot settings. Tabular results are in percent.}
    \large
	\resizebox{0.85\linewidth}{!}{
		\begin{tabular}{c|cc|cc|cc|cc}
			\toprule
			\multirow{3}{*}{Model} & \multicolumn{2}{c|}{{\textbf{3-shot}}} & \multicolumn{2}{c|}{{\textbf{5-shot}}} & \multicolumn{2}{c|}{{\textbf{10-shot}}} & \multicolumn{2}{c}{{\textbf{20-shot}}} \\
			\cmidrule{2-9}
			& Accuracy & F1 & Accuracy & F1 & Accuracy & F1 & Accuracy & F1 \\ 

            \bottomrule
            \toprule
            \multicolumn{9}{c}{\textbf{Dataset: IntentChat}} \\
            \midrule
   
		SAID & 52.73 & 58.49 & 63.75 & 67.80 & 69.18 & 72.54 & \textbf{74.05} & \textbf{76.44} \\
		\midrule
		+ Linear & \underline{65.86} & \underline{68.84} & \underline{71.58} & \underline{74.14} & 70.33 & 74.09 & 69.25 & 73.66 \\
            + MLP & 61.75 & 67.17 & 68.28 & 72.60 & \underline{70.51} & \underline{74.67} & 71.24 & 75.40 \\
            + QueryAdapt & \textbf{67.62} & \textbf{69.89} & \textbf{73.07} & \textbf{75.92} & \textbf{75.03} & \textbf{77.88} & \underline{72.08} & \underline{75.76} \\ 
            \midrule
            Improvement & \textbf{28.24\%} & \textbf{19.49\%} & \textbf{14.62\%} & \textbf{11.98\%} & \textbf{8.46\%} & \textbf{7.36\%} & -2.66\% & -0.89\% \\

            \bottomrule
            \toprule
            \multicolumn{9}{c}{\textbf{Dataset: WildChat}} \\
            \midrule

            SAID & 50.03 & 53.05 & 51.20 & 54.73 & 57.17 & 63.01 & 56.95 & 62.25 \\
		\midrule
		+ Linear & \underline{53.65} & \underline{58.40} & 52.54 & 57.58 & \underline{58.04} & \underline{62.17} & \underline{60.11} & \underline{64.89} \\
            + MLP & 45.19 & 48.79 & \underline{56.58} & \underline{60.11} & 54.62 & 58.91 & 56.53 & 61.59\\
            + QueryAdapt & \textbf{56.58} & \textbf{61.64} & \textbf{58.08} & \textbf{63.75} & \textbf{60.11} & \textbf{64.01} & \textbf{65.18} & \textbf{68.13} \\ 
            \midrule
            Improvement & \textbf{13.09\%} & \textbf{16.19\%} & \textbf{13.44\%} & \textbf{16.48\%} & \textbf{5.14\%} & \textbf{1.59\%} & \textbf{14.45\%}	& \textbf{9.45\%} \\
            
            \bottomrule
	\end{tabular}}
	\label{tab:q6-1}
\end{table*}

\begin{table*}[h]
	\renewcommand{\arraystretch}{1.05} 
	\centering 
	\caption{Results for the QueryAdapt strategy on three different PLM-based backbones. Tabular results are in percent.}
    \large
	\resizebox{0.85\linewidth}{!}{
		\begin{tabular}{c|cc|cc|cc|cc}
			\toprule
			\multirow{3}{*}{Model} & \multicolumn{2}{c|}{{\textbf{3-shot}}} & \multicolumn{2}{c|}{{\textbf{5-shot}}} & \multicolumn{2}{c|}{{\textbf{10-shot}}} & \multicolumn{2}{c}{{\textbf{20-shot}}} \\
			\cmidrule{2-9}
			& Accuracy & F1 & Accuracy & F1 & Accuracy & F1 & Accuracy & F1  \\ 
   
            \bottomrule
            \toprule
            \multicolumn{9}{c}{\textbf{Dataset: IntentChat}} \\
            \midrule
            
		SAID (BERT) & 52.73 & 58.49 & 63.75 & 67.80 & 69.18 & 72.54 & 74.05 & 76.44 \\
            + QueryAdapt & \textbf{67.62} & \textbf{69.89} & \textbf{73.07} & \textbf{75.92} & \textbf{75.03} & \textbf{77.88} & 72.08 & 75.76 \\
            Improvement & \textbf{28.24\%} & \textbf{19.49\%} & \textbf{14.62\%} & \textbf{11.98\%} & \textbf{8.46\%} & \textbf{7.36\%} & -2.66\% & -0.89\%\\

            \midrule
            SAID (DistilBERT) & 50.17 & 53.41 & 56.24 & 60.38 & 59.66 & 63.87 & 65.94 & 69.20 \\ 
            + QueryAdapt & \textbf{60.76} & \textbf{63.81} & \textbf{57.12} & \textbf{62.34} & \textbf{60.01} & 62.93 & 63.22 & 67.25 \\ 
            Improvement & \textbf{21.11\%} & \textbf{19.47\%} & \textbf{1.56\%} & \textbf{3.25\%} & \textbf{0.59\%} & -1.47\% & -4.12\% & -2.82\% \\

            \midrule
            SAID (ALBERT) & 49.51 & 50.36 & 50.62 & 54.83 & 61.27 & 64.22 & 57.32 & 61.32 \\
            + QueryAdapt & \textbf{57.98} & \textbf{61.49} & \textbf{56.47} & \textbf{60.22} & 58.27 & 62.76 & \textbf{62.65} & \textbf{66.65} \\ 
            Improvement & \textbf{17.11\%} & \textbf{22.10\%} & \textbf{11.56\%} & \textbf{9.83\%} & -4.90\% & -2.27\% & \textbf{9.30\%} & \textbf{8.69\%} \\
            
            \bottomrule
            \toprule
            \multicolumn{9}{c}{\textbf{Dataset: WildChat}} \\
            \midrule
            
		SAID (BERT) & 50.03 & 53.05 & 51.20 & 54.73 & 57.17 & 63.01 & 56.95 & 62.25 \\
            + QueryAdapt & \textbf{56.58} & \textbf{61.64} & \textbf{58.08} & \textbf{63.75} & \textbf{60.11} & \textbf{64.01} & \textbf{65.18} & \textbf{68.13} \\
            Improvement & \textbf{13.09\%} & \textbf{16.19\%} & \textbf{13.44\%} & \textbf{16.48\%} & \textbf{5.14\%} & \textbf{1.59\%} & \textbf{14.45\%}	& \textbf{9.45\%} \\

            \midrule
            SAID (DistilBERT) & 34.38 & 33.29 & 45.90 & 50.14 & 50.43 & 54.88 & 49.48 & 53.83 \\ 
            + QueryAdapt & \textbf{47.52} & \textbf{48.89} & \textbf{47.41} & \textbf{50.78} & \textbf{50.84} & \textbf{56.86} & \textbf{56.39} & \textbf{62.08} \\ 
            Improvement & \textbf{38.22\%} & \textbf{46.86\%} & \textbf{3.29\%} & \textbf{1.28\%} & \textbf{0.81\%} & \textbf{3.61\%} & \textbf{13.97\%} & \textbf{15.33\%} \\

            \midrule
            SAID (ALBERT) & 43.30 & 48.47 & 52.03 & 56.97 & 58.75 & 62.89 & 59.03 & 64.20 \\
            + QueryAdapt & 42.42 & 46.35 & \textbf{53.12} & \textbf{57.68} & \textbf{60.76} & \textbf{65.39} & \textbf{62.30} & \textbf{65.86} \\ 
            Improvement & -2.03\% & -4.37\% & \textbf{2.09\%} & \textbf{1.25\%} & \textbf{3.42\%} & \textbf{3.98\%} & \textbf{5.54\%} & \textbf{2.59\%} \\
            
            \bottomrule
	\end{tabular}}
	\label{tab:q6-2}
\end{table*}

\subsection{Efficiency Analysis (RQ5)} 
\label{sec:efficiency}
In this section, we compare the training and inference time of the proposed model against several baseline methods designed for few-shot intent detection, with a focus on efficiency analysis. Following existing studies, we evaluate and report the average training time per epoch \cite{liu2021beyond} and the average inference time per query (commonly referred to as inference latency). The experimental results using the IntentChat dataset are summarized in Table \ref{tab:effi}.

\noindent\textbf{Training efficiency.} The proposed models, SAID (BERT) and SAID (ALBERT), demonstrate competitive training efficiency compared to the baselines. With training times ranging from 1 to 15 seconds per epoch, these models are computationally efficient and well within acceptable limits, particularly given that training is conducted offline.

\noindent\textbf{Inference efficiency.} In terms of the more important inference latency, both SAID (BERT) and SAID (ALBERT) demonstrate comparable or even superior efficiency to the most efficient baseline CPFT, while providing significant improvements in terms of intent detection performance. This characteristic makes them ideal for online services that often require low-latency, real-time, and high-quality responses.

\subsection{Effectiveness of SAID (+QueryAdapt) (RQ6)} \label{sec:ex_rq6}
In this section, we investigate the explicit knowledge transfer strategy based on soft prompt generation. Building upon the SAID framework through task reformulation, we further propose to explicitly transfer relation tokens by generating intent-specific relation tokens explicitly during model fine-tuning and inference. In addition to the query-adaptive attention network (\textit{i.e.}, QueryAdapt), we also design two additional simple transfer strategies, namely Linear and MLP in this section. Specifically, the Linear strategy introduces two global weights, $\lambda_{qq}$ and $\lambda_{qa}$ defined in Equation (\ref{eq:relation_token_att}), which are shared across all user queries and optimized during model fine-tuning. On the other hand, the MLP strategy utilizes the query-query and query-answer relation tokens as input and generates intent-specific relation tokens through a feedforward deep neural network. The experimental results on the two datasets are summarized in Table \ref{tab:q6-1}.

\noindent\textbf{Effectiveness of the explicit knowledge transfer strategy.} In general, the proposed explicit knowledge transfer strategy via soft prompt generation can help to improve the overall model performance. As shown in Table \ref{tab:q6-1}, in most cases, the Linear, MLP, and QueryAdapt extensions significantly outperform their counterpart, the default model SAID. For example, SAID (+Linear) outperforms SAID by 24.90\% in terms of Accuracy and 17.70\% in terms of F1 score under the 3-shot setting on the IntentChat dataset. Similar findings can also be observed when comparing SAID (+MLP) and SAID (+QueryAdapt) with the default model SAID.

\noindent\textbf{Effectiveness of the query-adaptive attention network.} Compared with the simple transfer strategies (\textit{i.e.}, Linear and MLP), our proposed query-adaptive attention network achieves the best performance on two datasets across all the studied settings. The main difference between the Linear and QueryAdapt strategies lies in whether the learnable weights for generating intent-specific relation tokens are query dependent. As seen from Table \ref{tab:q6-1}, QueryAdapt achieves at least 3.88\% and 7\% higher Accuracy than Linear on the IntentChat and WildChat datasets, respectively. This result supports our motivation that different user queries may benefit from pretrained relations to varying degrees when inferring user intent. Meanwhile, we can find that Linear significantly outperforms the MLP strategy in most cases, suggesting that simply introducing non-linearity and additional parameters does not necessarily lead to improved model performance. 

\subsection{QueryAdapt on Various Backbones (RQ7)}
In this section, we evaluate the plug-and-play capability of the proposed explicit knowledge transfer strategy across different PLM-based backbones. Specifically, we adopt three widely used PLMs, including BERT, DistilBERT and ALBERT (as in Section~\ref{sec:rq2}). The experimental results on both datasets are reported in Table \ref{tab:q6-2}, demonstrating the performance improvements of the QueryAdapt empowered variant over the corresponding default SAID framework. These results show that the proposed QueryAdapt strategy consistently yields significant improvements across various PLMs that differ substantially in architecture and/or model size, under most settings.

\begin{figure}[th]
    \centering   \includegraphics[width=0.70\linewidth]{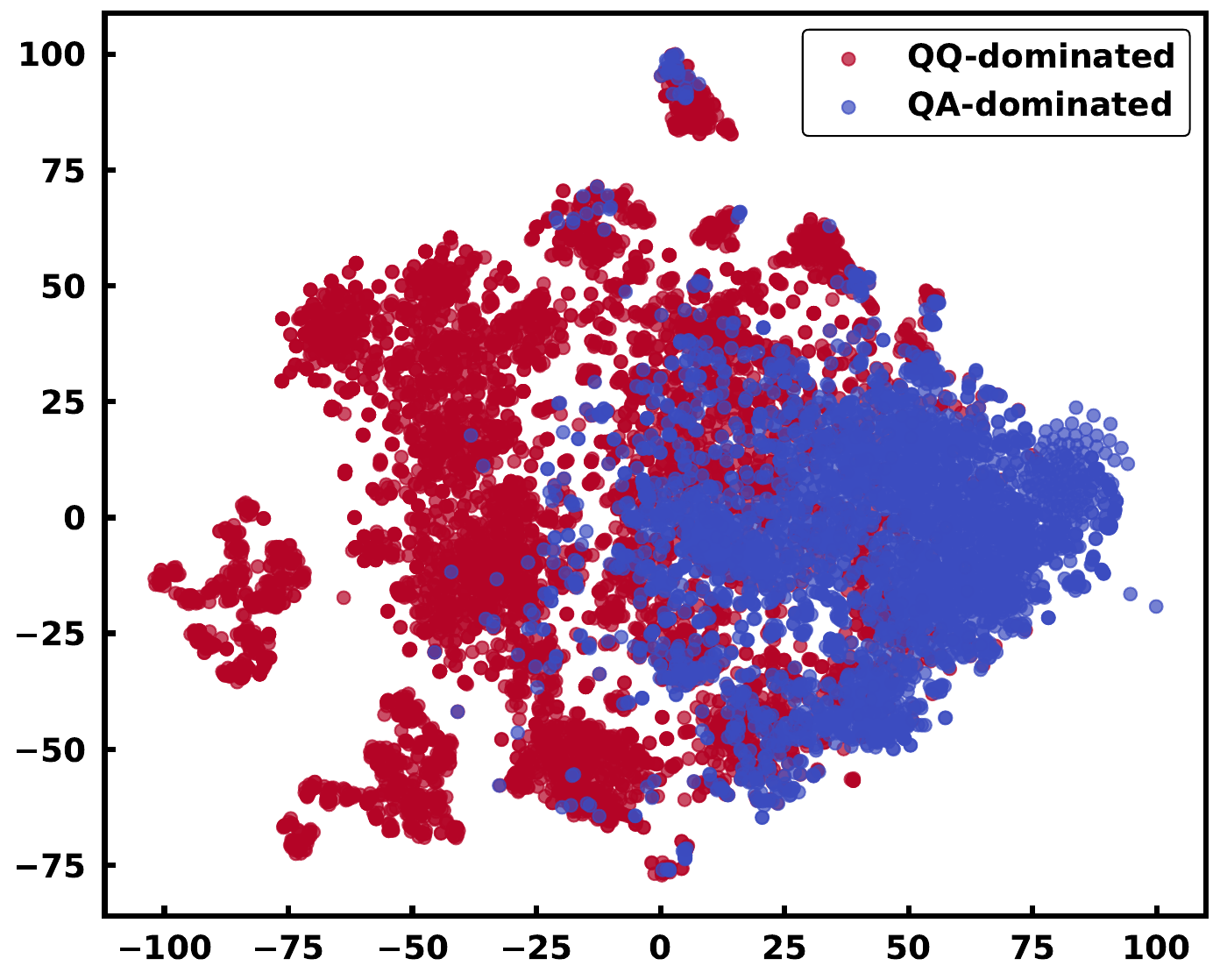}
    \vspace{-2mm}
    \caption{Query embedding visualizations on IntentChat.}
    \vspace{-2mm}
    \label{fig:query_tsne}
\end{figure}

\begin{figure}[t!]
	\centering
	\begin{subfigure}[b]{0.235\textwidth}
		\includegraphics[width=\textwidth]{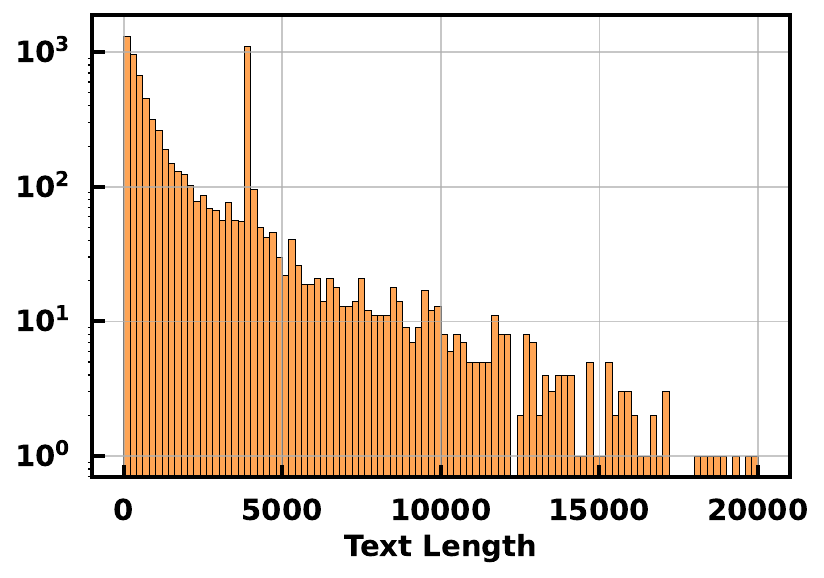}
		\caption{QQ-dominated}
		\label{fig:qq_dominate}
	\end{subfigure}
	\begin{subfigure}[b]{0.235\textwidth}
		\includegraphics[width=\textwidth]{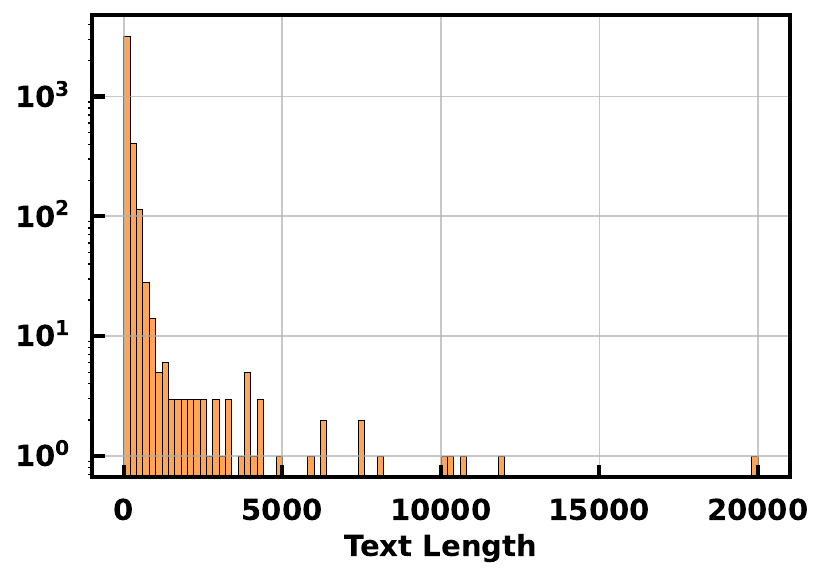}
		\caption{QA-dominated}
		\label{fig:qa_dominate}
	\end{subfigure}
    \vspace{-3mm}
	\caption{Comparison of query texts between QQ- and QA-Dominated categories on IntentChat.}
	\label{fig:att_case}
    \vspace{-3mm}
\end{figure}

\subsection{Case Study on Attention (RQ8)}
In this section, we conduct a case study to examine what the attention weights $\lambda_{qq}$ and $\lambda_{qa}$ have learned using the IntentChat dataset. First, we divide user queries in the test set into two categories: QQ-dominated, where the estimated $\lambda_{qq}$ for a given user query is greater than $\lambda_{qa}$, and QA-dominated, where the opposite holds. Then, we employ the t-SNE model \cite{visualizing2008} to project the query text embeddings, obtained from the fixed PLM encoder $f_{\text{plm}}(\cdot)$ in Equation (\ref{eq:relation_token_att}), into a two-dimensional space. The visualization results are shown in Figure~\ref{fig:query_tsne}. Each dot in the figure represents a user query, with its color indicating the corresponding category, \textit{i.e.}, QQ-dominated or QA-dominated. As shown in the figure, the two categories form distinct clusters with relatively clear boundaries.

We then compare the query texts between the two categories to better understand why certain user queries benefit more from query-query pretrained relations, while others do not. Specifically, we plot the text length distributions of user queries from both categories in Figure~\ref{fig:att_case}. The figure reveals that QQ-dominated queries (\textit{i.e.}, $\lambda_{qq}$ is greater) tend to be longer on average. This may be because longer queries often contain more diverse and potentially noisy information. In such cases, it becomes especially important to rely more on query refinement, which refers to how users refine and clarify the core purpose of their query (as encoded by the query-query relation tokens during pretraining), to help extract key information for accurate intent detection. 

\section{Conclusion}
\label{sec:conclusion}
In this paper, we propose a novel structure-aware pretraining framework, SAID, for the few-shot query intent detection problem. This framework unifies both textual and relational structure information inherent in conversational systems for model pretraining, marking the first attempt in this field to the best of our knowledge. SAID features a flexible relation-aware prompt and a structure-aware masked language modeling module for pretraining, along with prompt learning-based task reformulation for few-shot fine-tuning to facilitate effective transfer learning. 
Beyond the task reformulation via relation-aware prompt at the modeling paradigm level, we further introduce an enhanced model named SAID (+QueryAdapt) empowered by a query-adaptive attention network, which operates at the relation token level to enable more fine-grained knowledge transfer from pretraining to downstream few-shot intent detection. Empirical results show significant performance improvements over state-of-the-art methods. Furthermore, our framework can be seamlessly integrated with various backbone models, demonstrating its versatility and robustness.

\bibliographystyle{IEEEtran}
\bibliography{sample-base}

\end{document}